\documentclass{bmvc2k}

\newcommand{\abbrev}{PromptMoG}
\newcommand{\tmax}{\text{max}}
\usepackage{graphicx}
\usepackage{booktabs}
\usepackage{appendix}
\usepackage{multirow}
\usepackage{makecell}
\usepackage{arydshln}
\usepackage{aligned-overset}
\usepackage{amsmath} 
\usepackage{amsfonts}
\usepackage{amssymb}
\usepackage{colortbl}
\usepackage{xcolor}
\usepackage{enumitem}
\usepackage{etoc}
\usepackage{bm}
\usepackage{aliascnt}
\usepackage[ruled,linesnumbered]{algorithm2e}
\usepackage{bm}

\title{Diversifying Long Prompt Image Generation through Structured Prompt Embedding Space Sampling}

\addauthor{Bo-Kai Ruan}{bkruan.ee11@nycu.edu.tw}{1}
\addauthor{Teng-Fang Hsiao}{tfhsiao.ee13@nycu.edu.tw}{1}
\addauthor{Ling Lo}{linglo@cs.nthu.edu.tw}{2}
\addauthor{Yi-Lun Wu}{yilun.ee08@nycu.edu.tw}{1}
\addauthor{Hong-Han Shuai}{hhshuai@nycu.edu.tw}{1}

\addinstitution{
 Institute of Electrical and Computer Engineering\\
 National Yang Ming Chiao Tung University\\
 Hsinchu, Taiwan
}
\addinstitution{
 Department of Computer Science\\
 National Tsing Hua University\\
 Hsinchu, Taiwan
}

\runninghead{Ruan et al.}{PromptMoG}

\SetCommentSty{commentfont}
\usepackage{orcidlink}

\usepackage{tikz}

\usepackage[most]{tcolorbox}
\tcbset{
  promptstyle/.style={
    breakable,
    enhanced jigsaw,        % reliable page breaks
    colback=black!2,        % light background
    colframe=black!35,      % subtle frame
    boxrule=0.4pt,
    left=1em,right=1em,top=.6ex,bottom=.8ex,
    arc=1mm,                 % small rounding
    sharp corners=all,       % or comment this to keep rounding
    fonttitle=\bfseries,
    coltitle=black,
    colbacktitle=black!7,
  },
}
\newtcolorbox{prompt}[2][]{%
  promptstyle,
  title={#2}, % custom title text
  #1
}



\usepackage{amsthm}

\AtBeginEnvironment{appendices}{%
  \crefname{appendix}{Appendix}{Appendices}
  \crefalias{section}{appendix}
  \crefalias{subsection}{appendix}
}

\def\ie{\emph{i.e}\bmvaOneDot}
\def\eg{\emph{e.g}\bmvaOneDot}

\begin{document}

\maketitle

\begin{abstract}
Modern text-to-image models produce impressive visual results from richly specified prompts, yet their behavior under long prompts remains insufficiently understood. In this paper, we study a practical failure mode in which accumulated semantic constraints progressively suppress output variation, causing diversity to collapse even when many visual factors remain unspecified. We show that this phenomenon appears consistently across recent generation models as prompt length increases, and provide a theoretical motivation that connects long-prompt conditioning with reduced sampling entropy in the prompt embedding space. Based on this observation, we introduce \abbrev, a training-free approach that samples prompt embeddings from a Mixture-of-Gaussians distribution to restore generative flexibility while maintaining semantic fidelity. To support systematic evaluation, we further present LPD-Bench, a structured benchmark of long and semantically dense prompts for measuring both fidelity and diversity under compositional text conditioning. Extensive experiments on four large-scale diffusion models, including SD3.5-Large, Flux.1-Krea-Dev, CogView4, and Qwen-Image, show that \abbrev\ consistently improves diversity for long-prompt image generation. The code is publicly available at \url{https://github.com/basiclab/PromptMoG}.
\end{abstract}    
\section{Introduction}\label{sec:intro}

Recent text-to-image (T2I) systems have become practical engines for content creation across design ideation, advertising, media production, education, and interactive prototyping~\cite{podell2024sdxl,peebles2023scalable,zheng2024cogview,flux1kreadev2025,wu2025qwen,cai2025hidream}. In real-world applications, two core properties are crucial: \textit{fidelity} and \textit{diversity}. Fidelity ensures that generated images accurately reflect user intent, while diversity allows exploration of multiple plausible realizations of the same instruction. Importantly, a diverse set of outputs supports creative exploration, facilitates downstream selection and editing, and reduces the risk of stereotypical or repetitive renderings~\cite{xie2025sana,zhuo2025from}. Although modern diffusion models are less prone to mode collapse than earlier GANs~\cite{brock2018large,wu2021gradient,esser2021taming}, maintaining a balance between fidelity and diversity remains a central challenge. In particular, how this balance behaves under long prompts remains insufficiently explored.

\begin{figure}[t]
    \centering
    \includegraphics[width=0.8\linewidth]{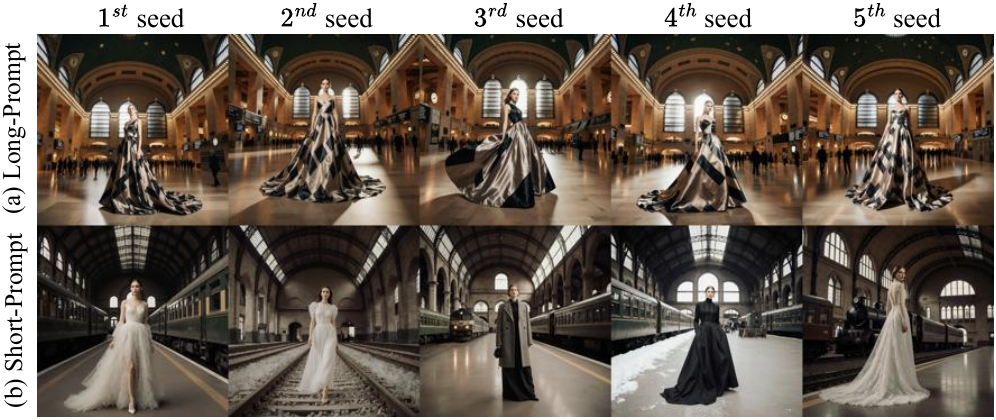}
    \caption{\textbf{Comparison of diversity generated with Flux.1-Krea-Dev.} Images are generated from different random seeds using (\textbf{a}) long prompts and (\textbf{b}) short prompts.}
    \label{fig:image_diversity_shown}
    \vspace{-10pt}
\end{figure}

In practical content-creation scenarios, prompts are rarely short descriptions. Instead, users often provide long, compositional prompts that specify multiple attributes, spatial relationships, lighting conditions, and stylistic cues. Such prompts introduce high semantic-density conditioning, where numerous constraints interact to guide image generation. While these detailed instructions improve semantic alignment, they also reshape the generative behavior of diffusion models. In particular, we observe that as prompt complexity increases, output diversity decreases markedly despite preserved prompt fidelity. \cref{fig:image_diversity_shown} illustrates this effect. When conditioned on a short prompt, diffusion models generate a wide range of variations across random seeds. However, when the prompt is expanded with specific elements such as soft daylight and a silk-shimmering gown, the generations converge to highly similar results. More importantly, this convergence also affects aspects that are \textit{not explicitly constrained}, such as background composition or hairstyle, indicating that long prompts can implicitly suppress variation beyond the specified content. This reduction in diversity limits the creative flexibility of modern T2I systems, where dominant semantic cues restrict alternative plausible interpretations.

Existing approaches for improving diversity in diffusion models fall into three categories: training-based, gradient-based, and sampling-based methods. Training-based approaches~\cite{miao2024training,datta2024prompt,dombrowski2025image} modify training objectives or architectures to encourage variability, but require retraining and substantial computational resources, limiting their applicability to modern T2I models. Gradient-based methods~\cite{askari2024improving,jalali2025sparke} optimize auxiliary objectives during inference, yet introduce significant computational and memory overhead, especially for high-resolution generation. Sampling-based approaches avoid retraining and are therefore more scalable. However, single-sampling methods~\cite{sadat2024cads} often rely on loosely constrained perturbations that may cause semantic drift, while joint-sampling methods~\cite{corso2024particle,morshed2025diverseflow,kirchhof2025shielded} require batch generation and additional optimization, leading to increased memory consumption. Moreover, these methods are primarily evaluated on short or moderately descriptive prompts with relatively sparse constraints. Consequently, they do not explicitly address the diversity degradation with long, compositionally dense prompts, nor provide scalable solutions for modern large-scale T2I systems. These limitations motivate the need for a training-free and scalable method that remains robust under dense compositional conditioning.

In this paper, we first analyze the effect of long prompts on diffusion models. As prompts become longer and compositionally dense, the conditioning signal increasingly concentrates around dominant semantic configurations, reducing the effective diversity of generated samples. Motivated by this observation, we develop a theoretical perspective that links diversity preservation to the sampling entropy of prompt embeddings. Guided by this principle, we propose \abbrev, a training-free and gradient-free method that restores generative diversity by increasing sampling entropy within the prompt embedding space. Specifically, \abbrev\ samples prompt embeddings from a Mixture-of-Gaussians distribution, enabling controlled exploration of semantically consistent variations without requiring model retraining. As a single-sampling strategy, our approach remains computationally efficient and scales naturally to modern large-scale T2I models. In addition, since our \abbrev\ operates in prompt embedding space, it is also compatible with autoregressive-based models.

Since the diversity behavior of T2I models under long compositional prompts remains underexplored, we introduce \underline{L}ong-\underline{P}rompt \underline{D}iversity \underline{Bench}mark (LPD-Bench), a structured benchmark designed for high semantic-density prompts. LPD-Bench draws from the photography taxonomy in Adobe Lightroom~\cite{adobe_photography_types} to construct 25 scene categories and 1,000 prompts (250–450 words each) that jointly specify semantic content, spatial layouts, and stylistic attributes. To quantify generative diversity, LPD-Bench adopts the Vendi Score~\cite{friedman2023the}, while a chunk-based evaluation protocol measures alignment between generated content and key prompt segments, enabling joint assessment of fidelity, diversity, and their trade-off. Experiments on several state-of-the-art T2I models, including SD3.5-Large~\cite{esser2024scaling}, Flux.1-Krea-Dev~\cite{flux1kreadev2025}, CogView4~\cite{zheng2024cogview}, and Qwen-Image~\cite{wu2025qwen}, show that \abbrev\ consistently improves diversity under long prompts while preserving semantic fidelity, outperforming heuristic baselines such as prompt chunking, prompt expansion~\cite{datta2024prompt}, and sampling-based approaches~\cite{sadat2024cads,morshed2025diverseflow}.
Our contributions can be summarized as follows:
\begin{itemize}
    \item We identify and analyze the diversity degradation phenomenon under long compositional prompts, showing that increasing semantic density induces a fidelity–diversity dilemma in modern diffusion-based text-to-image models.
    \item We propose \abbrev, a training- and gradient-free method that restores generative diversity by increasing sampling entropy in the prompt embedding space, improving diversity in diffusion models while also being compatible with autoregressive models.
    \item We introduce LPD-Bench, a benchmark for evaluating diversity and fidelity under long, detailed prompts, and show that \abbrev\ consistently improves diversity across large-scale T2I models while maintaining fidelity.
\end{itemize}
\section{Related Work}

\paragraph{\textbf{Text-to-Image Generation.}} Recent advances in T2I generation are largely driven by the success of latent diffusion models~\cite{rombach2022high,podell2024sdxl}, which learn to progressively denoise latent representations conditioned on textual descriptions. Diffusion transformers~\cite{peebles2023scalable,ma2024sit,yao2025diffusion} improve scalability through transformer architectures, where model capacity strongly correlates with visual quality. Rectified Flow models~\cite{liu2023flow} further reformulate the diffusion process into a deterministic formulation that accelerates inference while maintaining generation quality. Beyond visual quality, recent studies have advanced the semantic fidelity of the generated results. MM-DiT~\cite{esser2024scaling} enhances cross-modal integration in RF-based transformers by directly concatenating both image and text tokens for attention blocks, enabling bidirectional information flow that improves text comprehension. Additionally, large language models are increasingly utilized as text encoders~\cite{zheng2024cogview,cai2025hidream,wu2025qwen} to enhance prompt understanding, thereby strengthening the alignment between textual intention and visual realization. Although long, compositionally rich prompts provide detailed semantic cues that enhance fidelity, their effects in modern T2I systems remain insufficiently understood. In contrast, \cite{zhang2026the} studies the relationship between prompts and diversity only for prompts of length up to 100. Existing studies on long-prompt generation~\cite{wang2025tit} primarily evaluate alignment and quality, paying little attention to how these complex prompts influence generation diversity.

\vspace{-8pt}
\paragraph{\textbf{Diversity Enhancement for T2I Generation.}} 
Generation diversity refers to the range of distinct outputs a model produces from a given prompt, reflecting its ability to avoid repetitive or biased patterns. Existing methods for improving diffusion diversity fall into three categories. Training-based approaches modify objectives or architectures to encourage variability, \eg, DIG~\cite{miao2024training}, DiADM~\cite{dombrowski2025image}, and PE~\cite{datta2024prompt}. However, they require retraining, are computationally expensive, and scale poorly to large models. Gradient-based approaches introduce guidance during inference to update latents~\cite{dhariwal2021diffusion}, such as c-VSG~\cite{askari2024improving} and SPARKE~\cite{jalali2025sparke}, but they require backpropagation and incur substantial memory overhead (\eg, $>$ 80GB for Qwen-Image at $1024\times1024$ resolution). Sampling-based methods manipulate the inference process without retraining. CADS~\cite{sadat2024cads} perturbs conditioning features, while joint-sampling techniques~\cite{corso2024particle,morshed2025diverseflow,kirchhof2025shielded} optimize multiple latent trajectories jointly. Although computationally lighter, single-sampling methods often rely on loosely constrained perturbations that cause semantic drift under long prompts, whereas joint-sampling may over-optimize surrogate objectives and requires simultaneous batch generation, leading to substantial memory consumption as the number of seeds increases (\eg, $>$ 80GB for Qwen-Image with 6 samples at $1024\times1024$ resolution). Overall, existing techniques are primarily designed for short prompts, scale poorly to modern large models, and fail to address the fidelity-diversity tradeoff introduced by long prompts.
\vspace{-4pt}
\section{Methods}

We propose \abbrev, a training-free method that enhances generative diversity in long-prompt T2I models without compromising semantic fidelity. We begin with preliminaries on Rectified Flow (RF) models, which serve as the primary backbone for our method. Next, we analyze the diversity degradation that arises with increasing prompt length, showing how long prompts lead RF samplers toward similar generative results. Building on these observations, we develop a theoretical motivation that supports diverse sampling by enhancing prompt embeddings from an entropy-based perspective. Finally, we introduce our \abbrev\ and discuss the key design choices that enable effective diversity enhancement.

\subsection{Preliminary}\label{sec:preliminary}

\paragraph{\textbf{Rectified Flow.}}
An RF-based model~\cite{liu2023flow} aims to learn a \textit{linear} probability path between the clean data distribution $\mathbf{x}_0 \sim p_{\text{real}}$ and the Gaussian noise distribution $\mathbf{x}_{t_{\tmax}} \sim \mathcal{N}(\mathbf{0}, \mathbf{I})$:
\begin{equation*}
    \mathbf{x}_{t_k} = (1 - t_k)\mathbf{x}_0 + t_k\mathbf{x}_{t_{\tmax}}, \quad \forall \, t_k \in [0, t_\tmax],
\end{equation*}
by predicting the velocity field $\mathbf{v}_\theta(\mathbf{x}_{t_k}, t_k)$ defined as:
\vspace{-2pt}
\begin{equation}\label{eq:con_ode}
    d\mathbf{x}_{t_k} = \mathbf{v}_\theta(\mathbf{x}_{t_k}, {t_k})\,d{t_k}, \quad \forall \, {t_k} \in [0, t_{\tmax}].
\end{equation}
Since the velocity of a linear path is constant and equal to the difference between the clean and noisy samples, the training objective for $\mathbf{v}_\theta$ is formulated as:
\begin{equation}
    \mathcal{L} = 
    \mathbb{E}_{{t_k} \in [0, t_{\tmax}], 
    \mathbf{x}_0\sim p_{\text{real}},
    \mathbf{x}_{t_{\tmax}} \sim \mathcal{N}(\mathbf{0}, \mathbf{I}) }
    \left[ 
    \|(\mathbf{x}_0 - \mathbf{x}_{t_{\tmax}}) - \mathbf{v}_{\theta}(\mathbf{x}_{t_k}, t_k)\|^2_2 \right].
\end{equation}
\paragraph{\textbf{Sampling.}} Once the velocity model $\mathbf{v}_\theta$ is trained, sampling is performed by starting from the noise distribution ($\mathbf{x}_{t_{\tmax}} \sim \mathcal{N}(\mathbf{0}, \mathbf{I})$) and discretizing the Euler ODE from~\cref{eq:con_ode} as:
\begin{equation}\label{eq:sample}
    \mathbf{x}_{t_{k-1}} = \mathbf{x}_{t_k} + (t_{k-1}-t_k)\mathbf{v}_\theta(\mathbf{x}_{t_k}, t_k).
\end{equation}

\subsection{Diversity Degradation under Long Prompts}\label{sec:method_diverse}

To investigate the phenomenon of diversity degradation, we conduct a controlled analysis using long prompts sampled from LPD-Bench (see~\cref{sec:dataset} for details). For each prompt, we generate images across \textbf{6} random seeds and quantify sample diversity using the Vendi Score~\cite{friedman2023the}.
We compare the diversity of the original long prompts with shorter variants constructed from the first sentence and the first three sentences of each prompt. As shown in~\cref{fig:diversity_comparison}, longer prompts consistently yield lower Vendi Scores across all evaluated models, indicating that diversity decreases as prompt length increases. A qualitative example from Flux.1-Krea-Dev~\cite{flux1kreadev2025} is shown in~\cref{fig:image_diversity_shown}, and a finer-grained analysis across prompt-length buckets is provided in~\cref{sec:prompt_length_diversity}.

We further interpret the diversity degradation from an entropy perspective ($H$), where a lower entropy indicates that the generated samples are more predictable and exhibit lower diversity. In particular, longer prompts typically impose more constraints on the generation process, thereby reducing uncertainty and leading to less diverse outputs. For instance, specifying style, camera angle, or composition restricts the feasible region of possible outputs and makes the results more predictable. Formally, let $\mathbf{X}$ be a random variable representing the sample space for generated samples $\mathbf{x}$, and let $c_{s}$ and $c_{l}$ denote the short and long prompts. Then:
\begin{equation}
    H(\mathbf{X}\mid c_{l}) \le H(\mathbf{X}\mid c_{s}),
\end{equation}
where the inequality is strict when the additional details in the long prompt provide informative constraints on $\mathbf{X}$. The derivation is provided in~\cref{sec:proof_long_short}

\subsection{Prompt Reformulation}\label{sec:prompt_reformulation}

Longer prompts typically include more detailed descriptions, thereby constraining the feasible space and reducing output diversity. To mitigate this determinism, we propose to increase the stochasticity of the conditioning process by introducing multiple reformulated prompts during generation. Intuitively, using a set of prompt variants adds ``modes'' to the conditioning space, relaxing the constraints imposed by a single prompt and expanding the entropy of the resulting latent distribution.

Formally, we define a set of reformulated prompts $\mathcal{C}_{\text{ref}}$, obtained through transformations from the original prompt $c$. During inference, the conditional distribution of the sample random variable $\mathbf{X}$ can be expressed with the reformulated prompts through the law of total probability as a weighted mixture:
\vspace{-1pt}
\begin{equation}
    p(\mathbf{X}\mid c)
    = \sum_{c' \in \mathcal{C}_{\text{ref}}}
      p(\mathbf{X}\mid c')\,p(c'\mid c).
    \vspace{-2pt}
\end{equation}
We aim to enhance generation diversity by constructing a valid reformulated prompt set $\mathcal{C}_{\text{ref}}$. The entropy of the resulting distribution can be expressed as (see~\cref{sec:proof_prompt_reformulate}):
\vspace{-2pt}
\begin{equation}
    H_n(\mathbf{X}\mid c)
    = \sum_{c' \in \mathcal{C}_{\text{ref}}}
    p(c'\mid c)\,H(\mathbf{X}\mid c')
    + I(\mathbf{X}; \mathbf{C}_{\text{ref}}\mid c),
    \vspace{-3pt}
\end{equation}
where $H_n$ denotes the entropy computed under the cardinality $|\mathcal{C}_{\text{ref}}| = n$, $\mathbf{C}_{\text{ref}}$ is the random variable representing the reformulated prompt, and $I(\cdot)$ denotes the conditional mutual information between the generated samples and the reformulated prompts. To ensure that $\mathcal{C}_{\text{ref}}$ is non-trivial, it is essential that $H_n$ forms a monotonically increasing sequence with respect to $n$. In other words, by adding more prompts, we can progressively enhance the diversity.

However, the monotonicity of $H_n$ with respect to $n$ does not generally hold without additional structure. In the following setting: \textbf{(1)} all conditional entropies $H(\mathbf{X}\mid c')$ are identical (denoted $h$), \textbf{(2)} each conditional distribution $p(\mathbf{X}\mid c')$ has disjoint support, and \textbf{(3)} the mixture weights are uniform, $p(c'\mid c)=1/n$, we obtain a monotonic entropy increase as the number of prompt variants grows (the full proof is provided in~\cref{sec:proof_increasing}):
\vspace{-3pt}
\begin{equation}\label{eq:increase}
    H_{n+1} - H_n
    = \log(n+1) - \log n > 0.
    \vspace{-2pt}
\end{equation}

\begin{figure}[t]
    \centering

    \begin{minipage}[t]{0.49\linewidth}
        \centering
        \includegraphics[width=\linewidth]{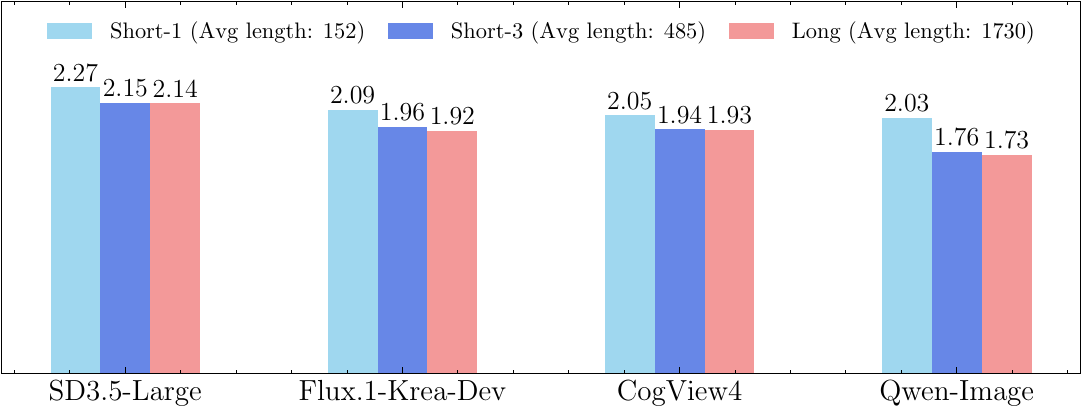}
        \captionof{figure}{\textbf{Comparison of diversity across different prompt lengths with Vendi score.} From left to right: using the first sentence, the first three sentences, and all sentences from each long prompt.}
        \label{fig:diversity_comparison}
    \end{minipage}\hfill
    \begin{minipage}[t]{0.49\linewidth}
        \centering
        \includegraphics[width=\linewidth]{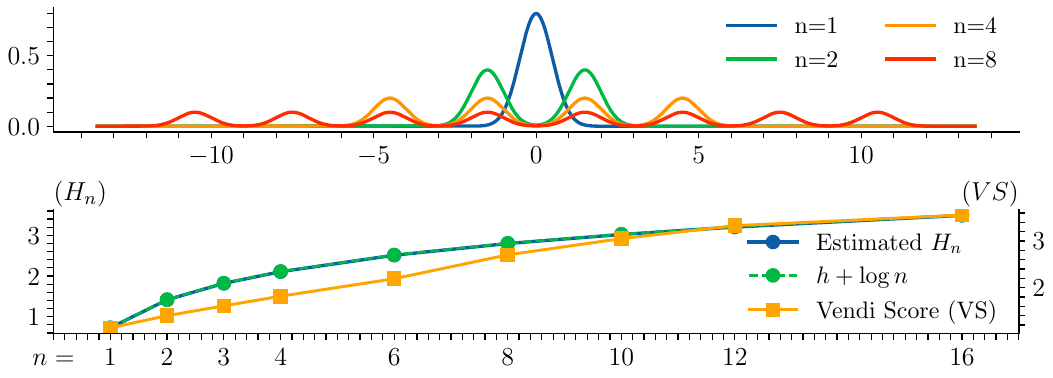}
        \captionof{figure}{\textbf{Illustration of the toy example.} (\textbf{Top}) 1D Mixture-of-Gaussians for varying $n$.
        (\textbf{Bottom}) Estimated entropy $H_n$ alongside the theoretical curve $h+\log n$, together with the Vendi Score.}
        \label{fig:toy_example}
    \end{minipage}

    \vspace{-10pt}
\end{figure}

\vspace{-10pt}
\paragraph{\textbf{Toy example.}}
We model the reformulated set as $n$ uniformly weighted 1D Gaussians,
$p_n(x)=\tfrac{1}{n}\sum_{k=1}^n \mathcal{N}(x;\mu_k,\sigma^2)$, with identical variance $\sigma^2$ and means spaced by $\Delta=6\sigma$ to approximate disjoint support, where each Gaussian represents a reformulated prompt.\footnote{In practice, the Gaussians are effectively disjoint when the $\boldsymbol{\mu}_i$ are sufficiently separated. Empirical validation is provided in~\cref{sec:disjointness}.}
We numerically estimate the entropy $H_n=-\!\int p_n(x)\log p_n(x)\,dx$ and compare it with the theoretical value $h+\log n$, where $h=\tfrac{1}{2}\log(2\pi e\sigma^2)$. As shown in~\cref{fig:toy_example}, the estimated entropy closely matches the theory and increases monotonically with $n$. The additional modes also spread probability mass across the real line, increasing sample diversity, consistent with the rising Vendi Score.

\subsection{\abbrev}\label{sec:design_decisions}

Motivated by the toy example, we adopt a simple yet effective MoG-based sampling strategy, namely \abbrev. Since the token space is discrete and semantically similar tokens are not necessarily positioned close to each other, we instead operate in the embedding space, where embeddings with similar meanings tend to lie in close proximity. Formally, we denote the prompt embedding of $c$ as $\boldsymbol{e}_c = E(c)$, where $E(\cdot)$ is the text encoder. To construct the MoG, we place the reformulated prompt embeddings $\{\boldsymbol{e}_{c'_i}\}_{i=1}^n$ so that the resulting mixture approximates the monotonicity conditions discussed above, thereby encouraging diversity to increase as the number of reformulated prompts $n$ grows. Once the MoG is defined, we sample a target embedding $\boldsymbol{e}_{c'}$ from the mixture,
\ie, $\boldsymbol{e}_{c'} \sim \frac{1}{n}\sum_i \mathcal{N}(\boldsymbol{e}_{c'_i}, \sigma^2 \mathbf{I})$,
and perform denoising conditioned on $\boldsymbol{e}_{c'}$. Below, we describe how to construct the reformulated prompt set to approximate the conditions that encourage entropy growth in~\cref{eq:increase}.
\vspace{-8pt}
\paragraph{\textbf{Identical Entropy.}}
To ensure that all conditional entropies $H(\mathbf{X}\mid c'_i)$ are similar, we adopt an effective strategy to make each reformulated prompt $c'_i$ close to the original prompt $c$. This design offers two advantages:  
\textbf{(1)} each $c'_i$ retains similar semantics, resulting in comparable effects on $\mathbf{x}$ during sampling and thus similar entropy values; and  
\textbf{(2)} the reformulated prompts stay semantically aligned with the original prompt, preventing unintended deviations in meaning. Since every $c'_i$ is derived from $c$, we define a feasible region as:
\vspace{-2pt}
\begin{equation}
    \|\boldsymbol{e}_c - \boldsymbol{e}_{c'_i}\|_2 \le \gamma_{\text{euc}}, 
    \quad \forall\, i \in \{1,\ldots,n\},
    \vspace{-2pt}
\end{equation}
where $\gamma_{\text{euc}}$ is a predefined similarity threshold controlling the semantic proximity between $c$ and its reformulations.  
In practice, since $\gamma_{\text{euc}}$ depends on the embedding dimension and is often difficult to determine directly, we first define a cosine similarity measure $\gamma_{\text{sim}}$ and then convert it to the corresponding Euclidean distance:
\vspace{-2pt}
\begin{equation}\label{eq:sim_to_euc}
    \gamma_{\text{euc}} = \|\boldsymbol{e}_c\|_2 \sqrt{2(1 - \gamma_{\text{sim}})},
    \vspace{-2pt}
\end{equation}
where we assume that $\{\boldsymbol{e}_{c'_i}\}$ and $\boldsymbol{e}_c$ have identical $\ell_2$ norms, as validated in~\cref{sec:norm}.

\vspace{-8pt}
\paragraph{\textbf{Disjointness and Uniform Probability.}}
Defining the feasible region $\mathcal{C}_{\text{ref}}$ as the Euclidean surface at distance $\gamma_{\text{euc}}$ from $\boldsymbol{e}_c$ forms a hypersphere in the embedding space. To ensure uniform mixture probabilities, the reformulated prompt embeddings $\{\boldsymbol{e}_{c'_i}\}$ are distributed \textbf{uniformly} on this hypersphere such that each embedding is equidistant from the others. This can be achieved using a \emph{regular simplex} construction, which guarantees maximal and uniform pairwise angular separation:
\vspace{-3pt}
\begin{equation}
    \boldsymbol{e}_{c'_i}^\top \boldsymbol{e}_{c'_j} 
    = -\,\gamma_{\text{euc}}^2 / (n-1),
    \quad \forall\, i \ne j.
    \vspace{-2pt}
\end{equation}
Since a regular simplex defines a fixed configuration for a given $n$, a rotation matrix can be applied to generate multiple uniform arrangements on the hyperspherical surface. Further details are provided in~\cref{sec:proof_simplex}. Although the simplex construction requires $n \le d+1$ in the $d$-dimensional space, prompt embeddings typically have high dimensionality (\eg, $d=1024$), so this constraint is not limiting in practice, and we find that a small $n$ (around $50$) already saturates the space. After using simplex construction to locate the reformulated prompts, these prompts, positioned on the hypersphere surface, serve as the centers of a MoG in the embedding space. To maintain disjointness among mixture components, we control the Gaussian variance $\sigma^2$ such that adjacent distributions remain non-overlapping. In practice, we set $\sigma$ relative to the embedding distance $\gamma_{\text{euc}}$ and thus scale it proportionally to $\gamma_{\text{euc}} / \sqrt{d}$. This ensures that the component spread adapts to the prompt embedding scale.
\vspace{-8pt}
\paragraph{\textbf{Overall Generation Pipeline.}}
The final prompt embeddings are sampled from the constructed MoG, and the subsequent image generation follows the original formulation. Specifically, we sample prompt embeddings at the token level, while keeping any pooled embedding (if present) fixed and leaving special tokens unchanged. The complete \abbrev\ algorithm and visual illustration are provided in~\cref{sec:ill_algo}.
\vspace{-8pt}
\paragraph{\textbf{Comparison with Pure Gaussian Noise.}}
Our sampling strategy differs fundamentally from directly adding pure Gaussian noise to the original prompt embedding (\eg, CADS~\cite{sadat2024cads}). Pure Gaussian noise tends to concentrate samples around the original embedding center, which preserves semantic fidelity but provides limited diversity. In contrast, \abbrev\ constructs a ring-like sampling region (see~\cref{sec:ill_algo}) by placing Gaussian components around the boundary of the feasible ball while leaving the center region less sampled. Under the same Gaussian scale, this design encourages prompt embeddings to explore more diverse directions while remaining close enough to preserve the original semantics. As a result, \abbrev\ achieves better fidelity and diversity results than pure Gaussian perturbation, as further supported by the Pareto analysis in~\cref{sec:results}, where our approach attains higher diversity in the high-fidelity region.
\vspace{-10pt}
\paragraph{\textbf{Extending to Autoregressive Models.}} Since our framework only requires access to the conditional distribution $p(\mathbf{X}\mid c)$, it naturally extends to autoregressive models.
By treating the subsequent image tokens as a random variable $\mathbf{X}$ and the given prompt embedding tokens as $c$, we can directly apply \abbrev\ by sampling prompt embeddings and then generating tokens under the resulting conditioning. More discussion can be found in~\cref{sec:auto_benchmarking}.
\section{LPD-Bench}\label{sec:dataset}

To systematically study the fidelity and diversity characteristics in long-prompt generation, we introduce LPD-Bench, a benchmark designed to evaluate the generative diversity, semantic fidelity, and aesthetic quality under long and compositionally rich prompts. Existing benchmarks~\cite{huang2023t2i,ghosh2023geneval,hu2024ella,li2024genai,Urbanek_2024_CVPR} primarily focus on short or less detailed prompts, failing to capture the semantic density and compositional complexity of long-form instructions. Therefore, they do not fit the diversity degradation of long and dense prompts in our setting. LPD-Bench addresses this gap by enabling systematic analysis of generation quality.

\subsection{Benchmark Design}\label{sec:dataset_construct}

LPD-Bench includes prompts drawn from diverse categories to ensure broad generalizability. We define a broad set of topics to cover a wide range of visual domains. Specifically, we reference the photography taxonomy in Adobe Lightroom~\cite{adobe_photography_types} to obtain representative scene categories, and merge semantically similar entries to form \textbf{25} distinct topics. Following prior works~\cite{hu2024ella,park2025raretofrequent}, we employ an LLM to generate detailed long prompts for each topic. Each sample contains a 250–450 word description that spans three aspects: \textit{content}, \textit{spatial}, and \textit{stylistic} attributes, simulating how humans naturally depict complex scenes. After data cleaning, we retain \textbf{1000} high-quality prompts, with 40 per topic. Details of topics, generation instructions, and data cleaning are provided in~\cref{sec:dataset_details}.

\begin{figure}[t]
    \centering
    \includegraphics[width=\linewidth]{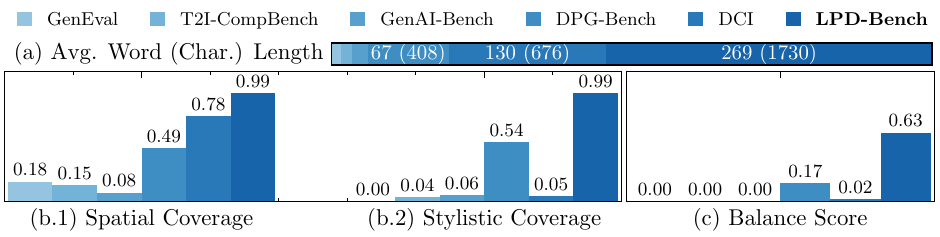}
    \vspace{-13pt}
    \caption{\textbf{Comparison of different benchmarks.} For clarity, the average length of the three smallest datasets is ignored.}
    \label{fig:dataset_lengths}
    \vspace{-10pt}
\end{figure}

As shown in~\cref{fig:dataset_lengths}, LPD-Bench provides prompts that are on average $2 \times$ longer in words and $2.5 \times $ longer in characters than those in existing benchmarks, making it well suited for evaluating long-prompt image generation. In addition, LPD-Bench offers the highest spatial and stylistic coverage, with a substantially larger proportion of prompts including spatial descriptions and stylistic cues that define the visual tone. Prior benchmarks rarely contain long or compositionally rich prompts, and therefore do not capture the complexities inherent in real-world descriptions. By design, most LPD-Bench samples naturally integrate both spatial and stylistic elements, which is reflected in a high \emph{balance score}\footnote{Entropy of spatial and stylistic tokens; higher indicates a more balanced composition (see~\cref{sec:data_statistics}).}, indicating a more even distribution of these two aspects within each prompt.

\subsection{Evaluation Protocol}\label{sec:dataset_eval}

To evaluate performance under long and compositionally rich prompts, LPD-Bench uses a multi-aspect protocol that measures diversity, semantic fidelity, and aesthetic quality. See~\cref{sec:overall_score} for details on computing the overall score.
\vspace{-8pt}
\paragraph{\textbf{Generative Diversity.}} We use Vendi Score (VS)~\cite{friedman2023the} to quantify diversity across $n$ outputs for each prompt using InceptionV3~\cite{szegedy2016rethinking} and DINOv3~\cite{simeoni2025dinov3} features to compute the similarity matrix. Unlike previous metrics that require reference datasets~\cite{heusel2017gans,binkowski2018demystifying,sajjadi2018assessing,naeem2020reliable}, Vendi Score directly evaluates diversity among generated samples, making it well-suited to our setting, where ground-truth distributions are unavailable.
\vspace{-10pt}
\paragraph{\textbf{Semantic Fidelity.}} We adopt a \textbf{chunk-based} evaluation protocol that considers \textit{content}, \textit{spatial}, and \textit{stylistic} fidelity. We segment each long prompt into these components, convert them into yes/no questions, and use \texttt{Qwen3-30B}~\cite{yang2025qwen3} to answer them conditioned on the generated image. We take the probability of a ``yes'' response as the VQA score~\cite{lin2024evaluating}. We also evaluate with \texttt{Gemma-4-31B-it}~\cite{team2026gemma} as an additional judge in~\cref{sec:add_vqa}.
\vspace{-10pt}
\paragraph{\textbf{Aesthetic.}} We estimate aesthetic quality using Aesthetic-Predictor-v2.5~\cite{schuhmann2022laion,as2024verb} to obtain Aesthetic Score (AS). Higher scores indicate stronger human preference and help reveal distortions or over-optimized artifacts.

\section{Experiments}\label{sec:experiments}

\paragraph{\textbf{Experimental Setup on Large-Scale Models.}}
We evaluate \abbrev\ on LPD-Bench using four large-scale T2I models: SD3.5-Large~\cite{esser2024scaling} (8B), Flux.1-Krea-Dev~\cite{flux1kreadev2025} (12B), CogView4~\cite{zheng2024cogview} (6B), and Qwen-Image~\cite{wu2025qwen} (20B). For each prompt, we generate images across \textbf{6 random seeds} to measure generation diversity. We compare \abbrev\ with representative sampling-based approaches, including the single-sampling method CADS~\cite{sadat2024cads} and the joint-sampling method DiverseFlow~\cite{morshed2025diverseflow}, since they do not require retraining or inference-time optimization and can be directly applied to large pretrained models. 
In addition, we include a prompt-chunking baseline that simulates shorter prompts, which typically produce higher diversity (see~\cref{sec:prompt_chunking}), and the prompt expansion baseline,  PE~\cite{datta2024prompt}. We also demonstrate how \abbrev\ can be applied to large-scale autoregressive models, including Show-o2~\cite{xie2025showo} and BLIP3o-NEXT~\cite{chen2025blip3o} (see~\cref{sec:auto_benchmarking}). All models are evaluated under their default settings, with hyperparameters provided in~\cref{sec:hyper}.
\vspace{-10pt}
\paragraph{\textbf{Experimental Setup on CC12M with LDM.}}
We conduct experiments on smaller diffusion models commonly used in diversity studies. Following SPELL~\cite{kirchhof2025shielded}, we evaluate LDM~\cite{rombach2022high} on the CC12M~\cite{changpinyo2021conceptual} dataset at $256\times256$ resolution, and compare with PG~\cite{corso2024particle} and IG~\cite{kynkanniemi2024applying}. We adopt the same evaluation protocol as SPELL, reporting precision~\cite{kynkaanniemi2019improved} and density~\cite{naeem2020reliable} for image quality, recall~\cite{kynkaanniemi2019improved}, coverage~\cite{naeem2020reliable}, and Vendi Score for diversity, and CLIP-T~\cite{radford2021learning} for prompt alignment.
\vspace{-10pt}
\paragraph{\textbf{Experimental Setup on MS-COCO with SD2.1.}}  We follow SPARKE~\cite{jalali2025sparke} and evaluate on the first 10,000 prompts from the MS-COCO~\cite{lin2014microsoft} 2014 validation set at $1024 \times 1024$ resolution. We also include c-VSG~\cite{askari2024improving} and CADS for comparison.  Following SPARKE, we report CLIP-T to assess text-image alignment, KID~\cite{binkowski2018demystifying} to measure distribution alignment, and conditional Vendi Score~\cite{friedman2023the,jalali2025information} to evaluate generative diversity, and in-batch similarity~\cite{corso2024particle} to quantify similarity among samples generated within the same prompt.

\begin{figure*}[t]
    \centering
    \includegraphics[width=\linewidth]{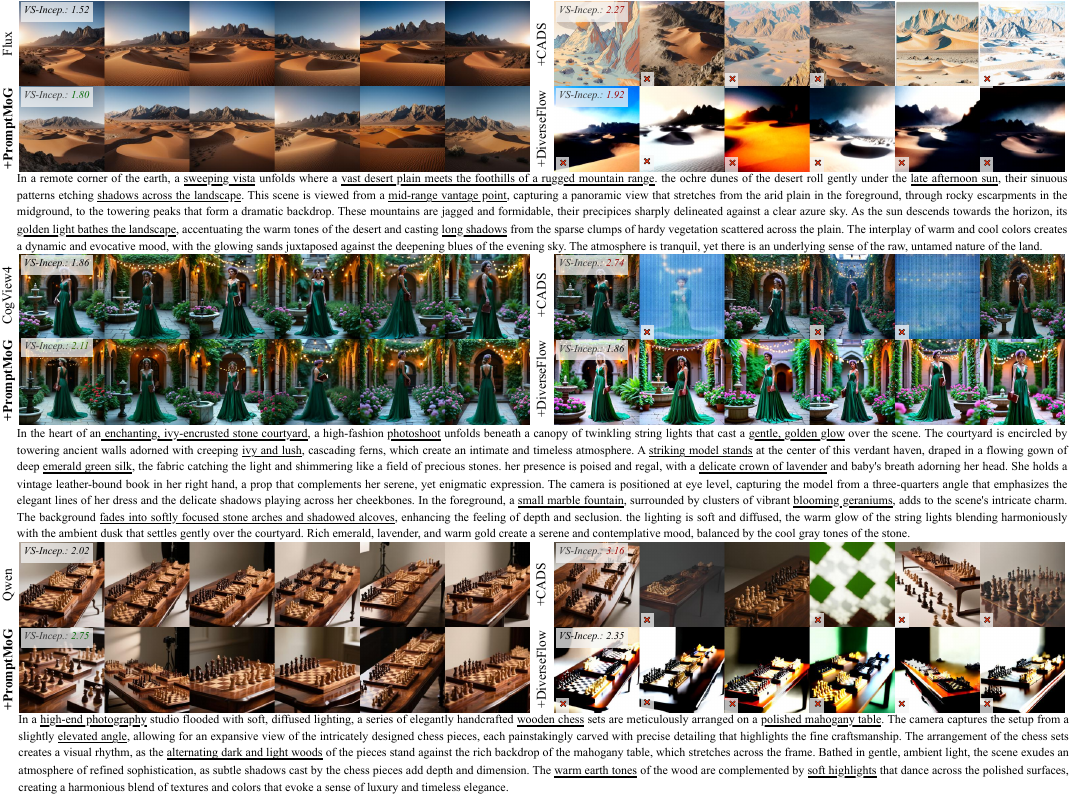}
    \vspace{-8pt}
    \caption{\textbf{Qualitative comparison with Flux.1-Krea-Dev (top), CogView4 (middle), and Qwen-Image (bottom).} Vendi Scores are shown at the top left, and outputs with visible artifacts are marked at the bottom left. (Full prompts in~\cref{sec:additional_comp}.)}
    \label{fig:qual}
    \vspace{-10pt}
\end{figure*}
{
\begin{table*}[t]
\centering
\resizebox{\linewidth}{!}{
\begin{tabular}{l*{8}{c}}
\toprule
\multirow{2.5}{*}{Method} & \multirow{2.5}{*}{Content-VQA} & \multirow{2.5}{*}{Spatial-VQA} & \multirow{2.5}{*}{Stylistic-VQA} & \multicolumn{5}{c}{\textbf{Average}} \\ \cmidrule(lr){5-9}
& & & & VQA & AS & VS-Incep.\footnotemark & VS-DINO & Overall $\downarrow$ \\
\midrule
% \rowcolor{NavyBlue!5}
SD3.5-Large~\cite{esser2024scaling} & 82.31 {\scriptsize $\pm$ 0.28} & 67.91 {\scriptsize $\pm$ 0.36} & 78.50 {\scriptsize $\pm$ 0.26} & 76.24 {\scriptsize $\pm$ 0.20} & 6.83 {\scriptsize $\pm$ 0.01} & 2.14 {\scriptsize $\pm$ 0.27} & 1.92 {\scriptsize $\pm$ 0.43} & - \\
% \rowcolor{NavyBlue!5}
+ Chunk & 62.95 {\scriptsize $\pm$ 0.37} & 56.43 {\scriptsize $\pm$ 0.35} & 69.28 {\scriptsize $\pm$ 0.46} & 62.89 {\scriptsize $\pm$ 0.26} & 6.69 {\scriptsize $\pm$ 0.01} & 2.26 {\scriptsize $\pm$ 0.32} & 2.13 {\scriptsize $\pm$ 0.53} & 0.83 \\
% \rowcolor{NavyBlue!5}
+ PE~\cite{datta2024prompt} & 34.20 {\scriptsize $\pm$ 1.00} & 33.93 {\scriptsize $\pm$ 0.89} & 46.24 {\scriptsize $\pm$ 0.87} & 38.12 {\scriptsize $\pm$ 0.86} & 5.45 {\scriptsize $\pm$ 0.05} & 3.17 {\scriptsize $\pm$ 0.39} & 4.19 {\scriptsize $\pm$ 0.84} & 0.53 \\
% \rowcolor{NavyBlue!5}
+ CADS~\cite{sadat2024cads} & 70.39 {\scriptsize $\pm$ 0.91} & 54.62 {\scriptsize $\pm$ 0.55} & 66.05 {\scriptsize $\pm$ 0.94} & 63.68 {\scriptsize $\pm$ 0.50} & 6.26 {\scriptsize $\pm$ 0.01} & 2.74 {\scriptsize $\pm$ 0.34} & 3.29 {\scriptsize $\pm$ 0.72} & 0.35 \\
% \rowcolor{NavyBlue!5}
+ DiverseFlow~\cite{morshed2025diverseflow} & 59.71 {\scriptsize $\pm$ 0.50} & 52.45 {\scriptsize $\pm$ 0.75} & 45.69 {\scriptsize $\pm$ 0.79} & 52.62 {\scriptsize $\pm$ 0.41} & 4.14 {\scriptsize $\pm$ 0.01} & 2.61 {\scriptsize $\pm$ 0.27} & 3.01 {\scriptsize $\pm$ 0.64} & 1.88 \\
\rowcolor{NavyBlue!5}
+ \abbrev & 78.66 {\scriptsize $\pm$ 0.19} & 65.59 {\scriptsize $\pm$ 0.25} & 76.75 {\scriptsize $\pm$ 0.18} & 73.67 {\scriptsize $\pm$ 0.06} & 6.80 {\scriptsize $\pm$ 0.01} & 2.20* {\scriptsize $\pm$ 0.29} & 2.04* {\scriptsize $\pm$ 0.48} & \textbf{0.31} \\

\noalign{\smallskip}
\hdashline
\noalign{\smallskip}

% \rowcolor{NavyBlue!5}
Flux.1-Krea-Dev~\cite{flux1kreadev2025} & 90.51 {\scriptsize $\pm$ 0.39} & 75.56 {\scriptsize $\pm$ 0.14} & 85.63 {\scriptsize $\pm$ 0.23} & 83.90 {\scriptsize $\pm$ 0.19} & 7.02 {\scriptsize $\pm$ 0.01} & 1.92 {\scriptsize $\pm$ 0.26} & 1.71 {\scriptsize $\pm$ 0.37} & - \\
% \rowcolor{NavyBlue!5}
+ Chunk & 68.28 {\scriptsize $\pm$ 0.34} & 62.35 {\scriptsize $\pm$ 0.34} & 74.08 {\scriptsize $\pm$ 0.29} & 68.24 {\scriptsize $\pm$ 0.14} & 6.89 {\scriptsize $\pm$ 0.01} & 2.08 {\scriptsize $\pm$ 0.34} & 2.12 {\scriptsize $\pm$ 0.53} & 0.50 \\
% \rowcolor{NavyBlue!5}
+ PE & 34.38 {\scriptsize $\pm$ 1.24} & 34.20 {\scriptsize $\pm$ 1.13} & 44.78 {\scriptsize $\pm$ 0.88} & 37.78 {\scriptsize $\pm$ 0.95} & 5.36 {\scriptsize $\pm$ 0.06} & 3.13 {\scriptsize $\pm$ 0.38} & 4.49 {\scriptsize $\pm$ 0.86} & 0.53 \\ 
% \rowcolor{NavyBlue!5}
+ CADS & 87.09 {\scriptsize $\pm$ 0.24} & 68.90 {\scriptsize $\pm$ 0.44} & 79.57 {\scriptsize $\pm$ 0.55} & 78.52 {\scriptsize $\pm$ 0.32} & 6.46 {\scriptsize $\pm$ 0.01} & 2.49 {\scriptsize $\pm$ 0.28} & 2.57 {\scriptsize $\pm$ 0.58} & 0.43 \\
% \rowcolor{NavyBlue!5}
+ DiverseFlow & 86.29 {\scriptsize $\pm$ 0.49} & 72.14 {\scriptsize $\pm$ 0.30} & 72.60 {\scriptsize $\pm$ 0.25} & 77.01 {\scriptsize $\pm$ 0.14} & 5.18 {\scriptsize $\pm$ 0.01} & 2.31 {\scriptsize $\pm$ 0.27} & 2.13 {\scriptsize $\pm$ 0.43} & 2.36 \\
\rowcolor{NavyBlue!5}
+ \abbrev & 88.26 {\scriptsize $\pm$ 0.29} & 73.75 {\scriptsize $\pm$ 0.20} & 84.82 {\scriptsize $\pm$ 0.20} & 82.27 {\scriptsize $\pm$ 0.12} & 7.02 {\scriptsize $\pm$ 0.00} & 2.01* {\scriptsize $\pm$ 0.30} & 1.86* {\scriptsize $\pm$ 0.43} & \textbf{0.07} \\

\noalign{\smallskip}
\hdashline
\noalign{\smallskip}

% \rowcolor{NavyBlue!5}
CogView4~\cite{zheng2024cogview} & 85.84 {\scriptsize $\pm$ 0.41} & 74.56 {\scriptsize $\pm$ 0.13} & 80.83 {\scriptsize $\pm$ 0.20} & 80.41 {\scriptsize $\pm$ 0.12} & 6.51 {\scriptsize $\pm$ 0.01} & 1.93 {\scriptsize $\pm$ 0.27} & 1.50 {\scriptsize $\pm$ 0.27} & - \\
% \rowcolor{NavyBlue!5}
+ Chunk & 26.32 {\scriptsize $\pm$ 0.30} & 35.88 {\scriptsize $\pm$ 0.40} & 51.39 {\scriptsize $\pm$ 0.41} & 37.87 {\scriptsize $\pm$ 0.25} & 5.57 {\scriptsize $\pm$ 0.01} & 2.31 {\scriptsize $\pm$ 0.44} & 2.27 {\scriptsize $\pm$ 0.75} & 1.19 \\
% \rowcolor{NavyBlue!5}
+ PE & 34.49 {\scriptsize $\pm$ 0.91} & 37.23 {\scriptsize $\pm$ 0.82} & 51.04 {\scriptsize $\pm$ 0.75} & 40.92 {\scriptsize $\pm$ 0.72} & 6.07 {\scriptsize $\pm$ 0.04} & 3.04 {\scriptsize $\pm$ 0.38} & 3.85 {\scriptsize $\pm$ 0.95} & \textbf{0.24} \\
% \rowcolor{NavyBlue!5}
+ CADS & 51.08 {\scriptsize $\pm$ 1.18} & 48.80 {\scriptsize $\pm$ 1.04} & 50.69 {\scriptsize $\pm$ 0.56} & 50.19 {\scriptsize $\pm$ 0.86} & 4.79 {\scriptsize $\pm$ 0.03} & 2.99 {\scriptsize $\pm$ 0.34} & 3.69 {\scriptsize $\pm$ 0.59} & 0.62 \\
% \rowcolor{NavyBlue!5}
+ DiverseFlow & 81.32 {\scriptsize $\pm$ 0.33} & 72.07 {\scriptsize $\pm$ 0.33} & 72.84 {\scriptsize $\pm$ 0.32} & 75.41 {\scriptsize $\pm$ 0.19} & 5.49 {\scriptsize $\pm$ 0.01} & 1.97 {\scriptsize $\pm$ 0.26} & 1.58 {\scriptsize $\pm$ 0.32} & 8.92 \\
\rowcolor{NavyBlue!5}
+ \abbrev & 81.18 {\scriptsize $\pm$ 0.43} & 70.94 {\scriptsize $\pm$ 0.43} & 79.09 {\scriptsize $\pm$ 0.39} & 77.07 {\scriptsize $\pm$ 0.32} & 6.49 {\scriptsize $\pm$ 0.00} & 2.02* {\scriptsize $\pm$ 0.31} & 1.63* {\scriptsize $\pm$ 0.36} & \textbf{0.24} \\

\noalign{\smallskip}
\hdashline
\noalign{\smallskip}

% \rowcolor{NavyBlue!5}
Qwen-Image~\cite{wu2025qwen} & 94.63 {\scriptsize $\pm$ 0.14} & 77.81 {\scriptsize $\pm$ 0.22} & 87.28 {\scriptsize $\pm$ 0.10} & 86.58 {\scriptsize $\pm$ 0.07} & 6.64 {\scriptsize $\pm$ 0.01} & 1.73 {\scriptsize $\pm$ 0.20} & 1.32 {\scriptsize $\pm$ 0.17} & - \\
% \rowcolor{NavyBlue!5}
+ Chunk & 65.87 {\scriptsize $\pm$ 0.36} & 61.85 {\scriptsize $\pm$ 0.39} & 76.80 {\scriptsize $\pm$ 0.21} & 68.17 {\scriptsize $\pm$ 0.16} & 6.66 {\scriptsize $\pm$ 0.01} & 1.83 {\scriptsize $\pm$ 0.24} & 1.48 {\scriptsize $\pm$ 0.30} & 0.63 \\
% \rowcolor{NavyBlue!5}
+ PE & 37.36 {\scriptsize $\pm$ 1.04} & 38.16 {\scriptsize $\pm$ 1.09} & 51.76 {\scriptsize $\pm$ 0.71} & 42.42 {\scriptsize $\pm$ 0.80} & 5.72 {\scriptsize $\pm$ 0.04} & 3.20 {\scriptsize $\pm$ 0.42} & 4.09 {\scriptsize $\pm$ 0.99} & 0.32 \\
% \rowcolor{NavyBlue!5}
+ CADS & 84.38 {\scriptsize $\pm$ 0.79} & 68.63 {\scriptsize $\pm$ 1.00} & 73.42 {\scriptsize $\pm$ 0.87} & 75.48 {\scriptsize $\pm$ 0.69} & 6.06 {\scriptsize $\pm$ 0.03} & 2.74 {\scriptsize $\pm$ 0.48} & 2.58 {\scriptsize $\pm$ 0.74} & 0.30 \\
% \rowcolor{NavyBlue!5}
+ DiverseFlow & 82.43 {\scriptsize $\pm$ 0.32} & 72.41 {\scriptsize $\pm$ 0.22} & 58.99 {\scriptsize $\pm$ 0.42} & 71.27 {\scriptsize $\pm$ 0.18} & 4.80 {\scriptsize $\pm$ 0.01} & 2.12 {\scriptsize $\pm$ 0.27} & 1.70 {\scriptsize $\pm$ 0.40} & 2.59 \\
\rowcolor{NavyBlue!5}
+ \abbrev & 92.31 {\scriptsize $\pm$ 0.20} & 75.76 {\scriptsize $\pm$ 0.42} & 86.42 {\scriptsize $\pm$ 0.19} & 84.83 {\scriptsize $\pm$ 0.14} & 6.63 {\scriptsize $\pm$ 0.01} & 1.88* {\scriptsize $\pm$ 0.25} & 1.49* {\scriptsize $\pm$ 0.27} & \textbf{0.09} \\
% \midrule
% \multicolumn{8}{c}{\textit{Closed-Source}} \\
% \midrule
% \rowcolor{gray!5}
% GPT-Image-1~\cite{openaigptimage1} & 92.70 {\scriptsize $\pm$ 0.23} & 77.73 {\scriptsize $\pm$ 0.26} & 86.13 {\scriptsize $\pm$ 0.16} & 85.52 {\scriptsize $\pm$ 0.12} & 6.64 {\scriptsize $\pm$ 0.00} & 1.75 {\scriptsize $\pm$ 0.21} & 1.43 {\scriptsize $\pm$ 0.21} \\
\bottomrule
\end{tabular}
}
\caption{
    \textbf{Quantitative results on LPD-Bench.} (*: $p\text{-value} \approx 10^{-5} \le 10^{-3}$, with Welch's t-test across prompts.)
}
\label{tab:lbp_results}
\vspace{-10pt}
\end{table*}
}

\subsection{Results}\label{sec:results}

\paragraph{\textbf{Qualitative Comparison.}} The results for Flux.1-Krea-Dev, CogView4, and Qwen-Image are shown in~\cref{fig:qual}, and additional comparisons are provided in~\cref{sec:additional_comp}. As illustrated, CADS introduces ill-constrained variations: although diversity increases, semantic fidelity collapses. For instance, the sky disappears and the style becomes distorted in the top example; a checkerboard artifact emerges in the middle example; and incorrect lighting or object placement appears in the bottom example. On the other hand, DiverseFlow suffers from over-optimization, producing visually degraded outputs with incorrect overall styles. In contrast, our constrained sampling generates diverse outputs while preserving the correct semantic content, as reflected by the Vendi Score and the visual results.
\vspace{-10pt}
\paragraph{\textbf{Quantitative Comparison on LPD-Bench.}} The quantitative results are presented in~\cref{tab:lbp_results}. For the chunking baseline, diversity increases, but semantic scores drop noticeably, as splitting and recombining prompt segments can disrupt the original meaning. A similar trend is observed for PE, which significantly degrades semantic fidelity, as expanding long prompts may require additional tuning. CADS also introduces severe distortions, consistent with the qualitative findings, as it adjusts perturbation scales across timesteps without enforcing semantic consistency. DiverseFlow exhibits over-optimization, leading to style degradation reflected in both the stylistic VQA and AS metrics. In contrast, our method introduces controlled variability, achieving higher diversity than the original model while consistently preserving semantic integrity and visual quality, as indicated by VQA and AS, with the best overall score. Improvements are also observed for autoregressive models in~\cref{sec:auto_benchmarking}.
\footnotetext{Since both Vendi scores are computed across all seeds for a prompt, the reported standard deviation reflects prompt variation rather than method uncertainty.}

\begin{figure}[t]
    \centering
    \includegraphics[width=\linewidth]{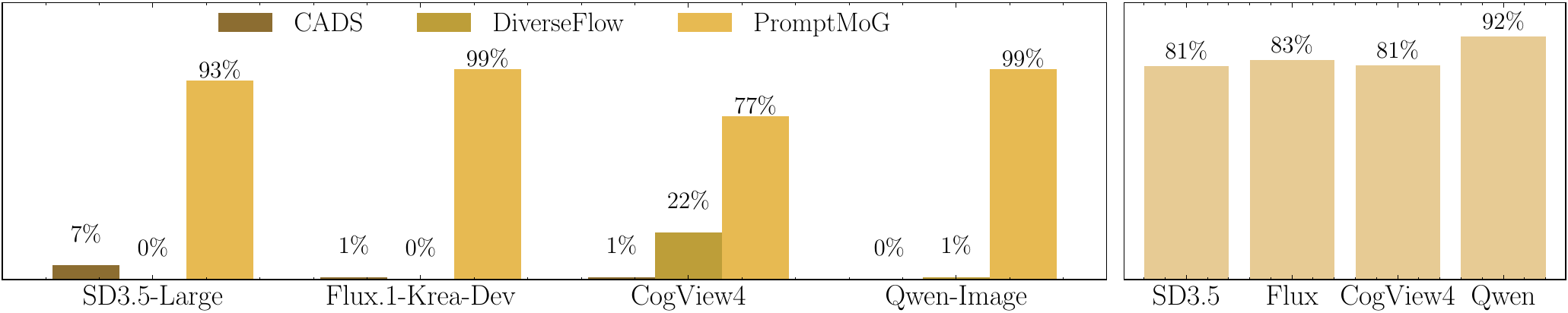}
    \vspace{-10pt}
    \caption{\textbf{User study.} \textbf{(Left)} Comparison of semantic preservation among diversity-enhancement methods. \textbf{(Right)} Diversity win-rate against the original model.}
    \vspace{-5pt}
    \label{fig:user_study}
\end{figure}
\begin{figure}[t]
    \centering
    \includegraphics[width=\linewidth]{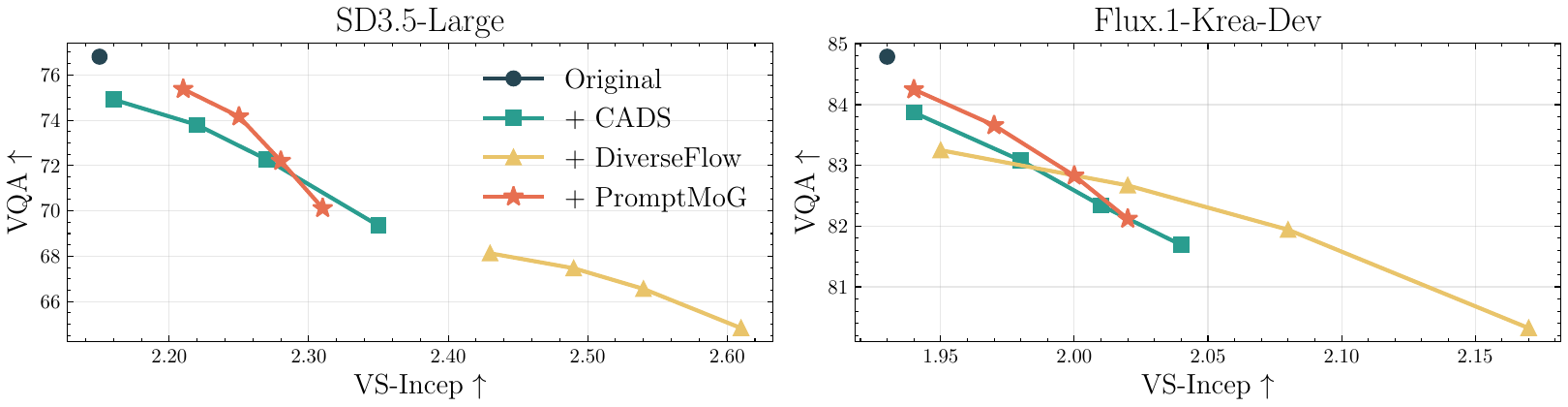}
    \vspace{-13pt}
    \caption{\textbf{Pareto plot on LPD-Bench.} We compare diversity and fidelity across different hyperparameter settings. Our method lies on the Pareto frontier in the high-fidelity region (upper-half), while low-fidelity points are less usable due to weak prompt alignment.}
    \vspace{-10pt}
    \label{fig:pareto}
\end{figure}
{
\renewcommand{\arraystretch}{0.95}
\setlength{\tabcolsep}{7pt}
\begin{table}[t]
\centering
\scriptsize
% \vspace{-5pt}
\resizebox{\linewidth}{!}{\begin{tabular}{lcccccccc}
\toprule
Method & Vendi Score & Recall & Coverage & Precision & Density & CLIP-T \\
\midrule
% \rowcolor{NavyBlue!5}
LDM~\cite{rombach2022high} & 2.527 & 0.237  & 0.446 & 0.558 & 0.768 & 27.789 \\
% \rowcolor{NavyBlue!5}
+ PG~\cite{corso2024particle} & 2.753 & 0.230 & 0.378 & 0.443 & 0.534 & 26.740 \\
% \rowcolor{NavyBlue!5}
+ IG~\cite{kynkanniemi2024applying} & \underline{3.208} & \textbf{0.471} & 0.421 & 0.483 & \underline{0.635} & 25.885 \\
% \rowcolor{NavyBlue!5}
+ SPELL~\cite{kirchhof2025shielded} & 2.998 & 0.370 & \underline{0.437} & \underline{0.500} & 0.631 & \underline{27.397} \\
\rowcolor{NavyBlue!5}
+ \abbrev & \textbf{3.912} & \underline{0.375} & \textbf{0.475} & \textbf{0.521} & \textbf{0.668} & \textbf{27.748} \\
\bottomrule
\end{tabular}}
\caption{\textbf{Quantitative results on CC12M using LDM.}}\label{tab:comp_joint}
\vspace{-10pt}
\end{table}
}

{
\renewcommand{\arraystretch}{0.95}
\setlength{\tabcolsep}{5pt}
\begin{table}[t]
  \scriptsize
  \centering
  \resizebox{\linewidth}{!}{\begin{tabular}{lccccc}
    \toprule
    Method & CLIP-T & KID ${\times 10^{-4}}$ & Cond-Vendi Score & Vendi Score & In-batch Sim. \\
    \midrule
    % \rowcolor{NavyBlue!5}
    SD2.1~\cite{rombach2022high} & 31.20 & 52.37 & 26.54 & 369.37 & 80.36 \\
    % \rowcolor{NavyBlue!5}
    + c-VSG~\cite{askari2024improving} & 28.75 & 56.25 & 27.91 & 376.48 & 79.82 \\
    % \rowcolor{NavyBlue!5}
    + CADS~\cite{sadat2024cads}  & 29.89 & 55.08 & 28.73 & 380.08 & 79.44 \\
    % \rowcolor{NavyBlue!5}
    + SPARKE~\cite{jalali2025sparke} & \underline{30.96} & \textbf{53.15} & \underline{32.57} & \underline{405.51} & \underline{75.68} \\
    \rowcolor{NavyBlue!5}
    + \abbrev & \textbf{31.83} & \underline{54.48} & \textbf{41.21} & \textbf{448.29} & \textbf{64.14}\\
    \bottomrule
  \end{tabular}}
\caption{\textbf{Quantitative results on MS-COCO using SD2.1.}}\label{tab:gradient-mscoco}
\vspace{-10pt}
\end{table}
}

\vspace{-10pt}
\paragraph{\textbf{User Study.}} To verify that the diversity gains align with human perception, we conduct a user study with 15 participants on 24 randomly sampled prompts as shown in~\cref{fig:user_study}. Results show that \abbrev\ is preferred over CADS and DiverseFlow in semantic preservation, while achieving a diversity win rate of over 80\% against the original models. This suggests that \abbrev\ improves diversity without relying on semantic drift.
\vspace{-10pt}
\paragraph{\textbf{Quantitative Comparison on CC12M.}} The results are shown in~\cref{tab:comp_joint}, where baseline results are adopted from SPELL. Our method significantly improves the Vendi Score while simultaneously increasing coverage, precision, density, and CLIP-T, indicating a more diverse yet semantically aligned, well-covered sample distribution and strong prompt alignment. In contrast, although IG achieves high recall, it substantially sacrifices coverage and precision and shows weaker semantic consistency with the prompt. Overall, our approach achieves a better balance between diversity and fidelity, even without joint sampling.
\vspace{-10pt}
\paragraph{\textbf{Quantitative Comparison on MS-COCO.}} As shown in~\cref{tab:gradient-mscoco}, our method achieves higher diversity, as reflected by both improved Vendi Scores and lower in-batch similarity, indicating reduced collapse within each prompt. Importantly, this increase in diversity does not substantially compromise fidelity, as evidenced by strong CLIP-T and the second-lowest KID, suggesting competitive distribution alignment. Overall, these results demonstrate that our approach remains competitive with gradient-based methods while maintaining a balanced tradeoff between diversity and quality.
\vspace{-10pt}
\paragraph{\textbf{Pareto Plot Across Different Settings.}} \cref{tab:lbp_results} reports results under the default hyperparameters. To assess robustness across different settings and avoid overestimating performance from a single configuration, we provide a Pareto plot in~\cref{fig:pareto} comparing our method with CADS and DiverseFlow. Due to computational cost, we restrict this analysis to sampling-based methods and follow the prompt setup in~\cref{sec:abla}. We use SD3.5-Large and Flux.1-Krea-Dev, the first two models in~\cref{tab:lbp_results}, as representative cases. In the high-fidelity region, corresponding to the upper half of the figure, our method achieves higher diversity while better preserving VQA performance, thereby forming the Pareto frontier. Although CADS or DiverseFlow may become Pareto-optimal in the low-fidelity region, corresponding to the lower half of the figure, this regime is less meaningful because the generated outputs no longer align well with the prompt. This observation is consistent with our discussion on Gaussian noise in~\cref{sec:design_decisions} and matches the comparison in~\cref{tab:lbp_results}.

\begin{figure}[t]
    \centering

    % Row 1 (unchanged)
    \begin{minipage}[t]{0.49\linewidth}
        \centering
        \includegraphics[width=\linewidth]{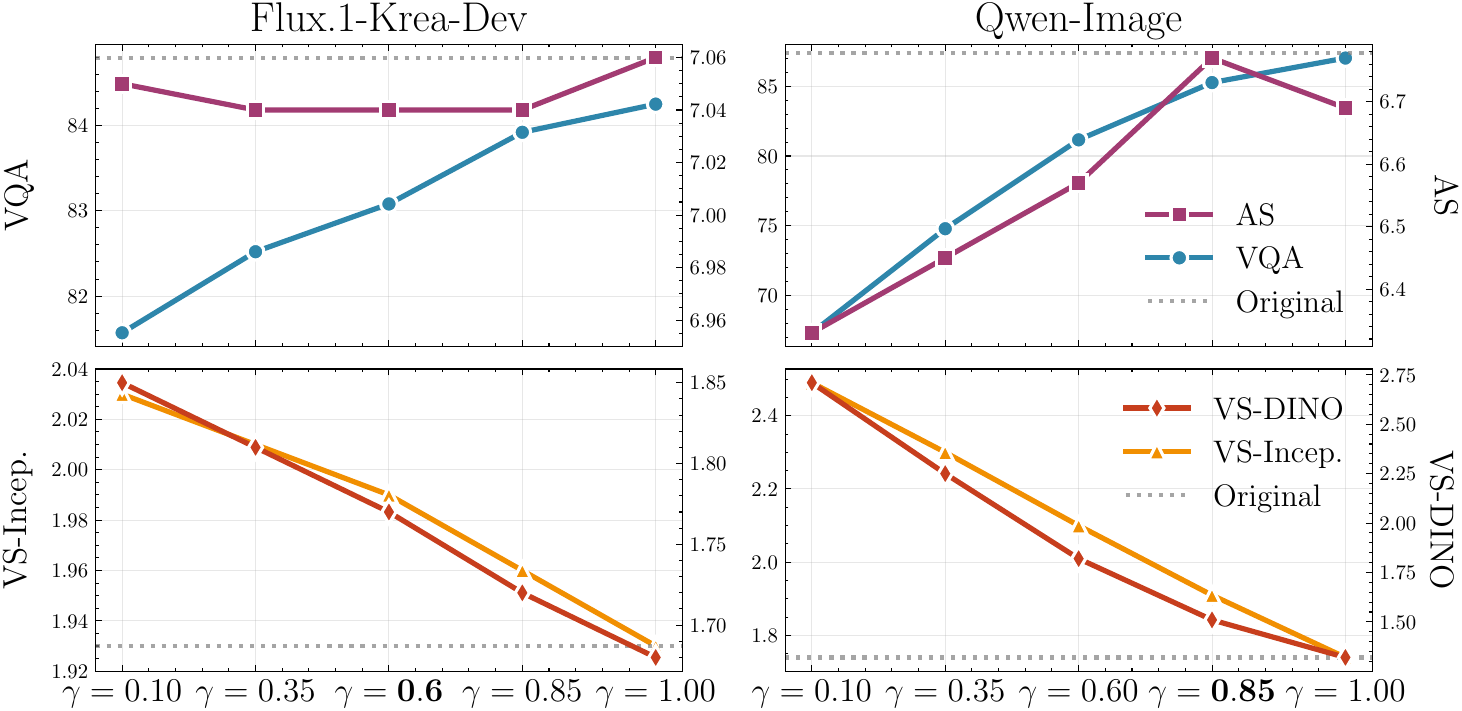}
        \vspace{-8.5pt}
        \captionof{figure}{\textbf{Comparison across different $\gamma_{\text{sim}}$.} The default setting is highlighted in \textbf{bold}. Dashed line: w/o. \abbrev.}
        \label{fig:abla_gamma}
    \end{minipage}
    \hfill
    \begin{minipage}[t]{0.49\linewidth}
        \centering
        \includegraphics[width=\linewidth]{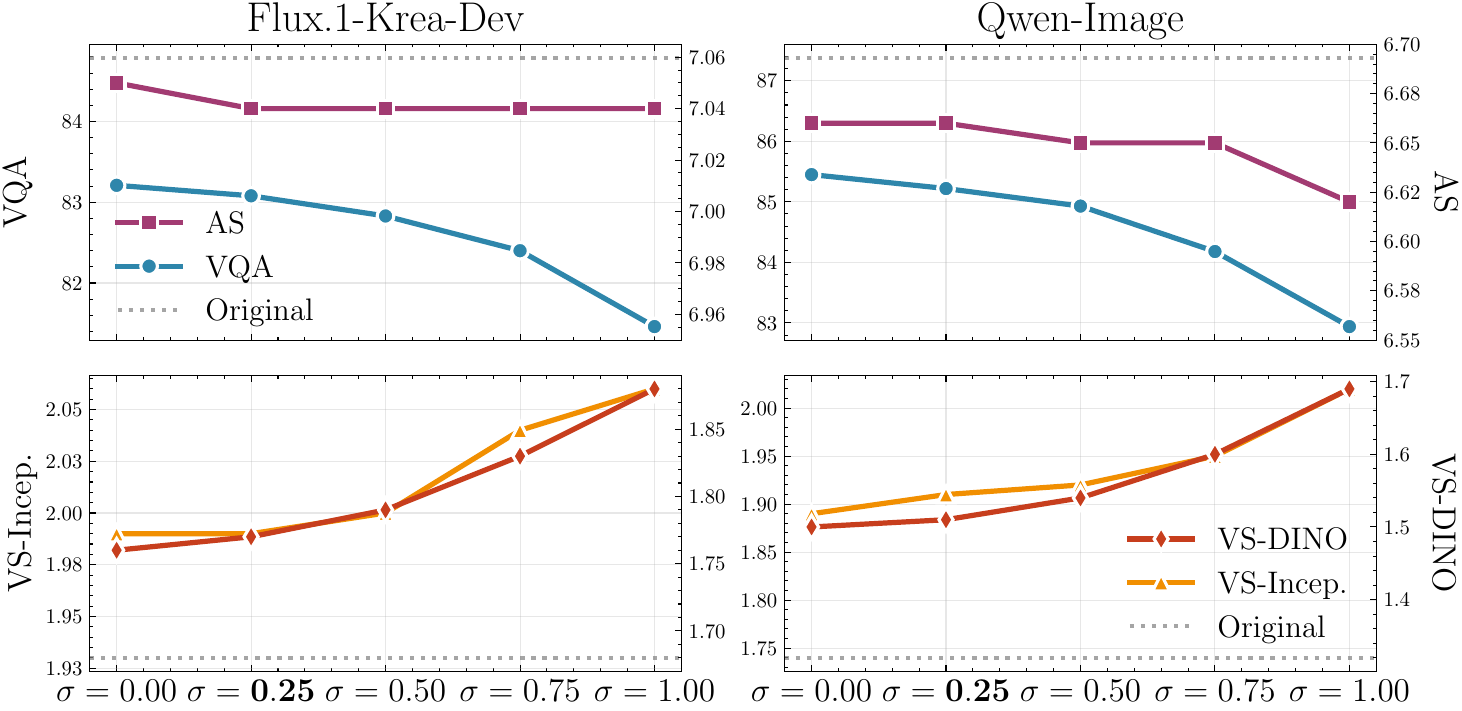}
        \vspace{-8.5pt}
        \captionof{figure}{\textbf{Comparison across different $\sigma$.} The default setting is highlighted in \textbf{bold}. Dashed line: w/o. \abbrev.}
        \label{fig:abla_sigma}
    \end{minipage}

    \vspace{3pt} % space between rows

    % Row 2 (left unchanged, right merged vertically)
    \begin{minipage}[t]{0.49\linewidth}
        \centering
        \includegraphics[width=\linewidth]{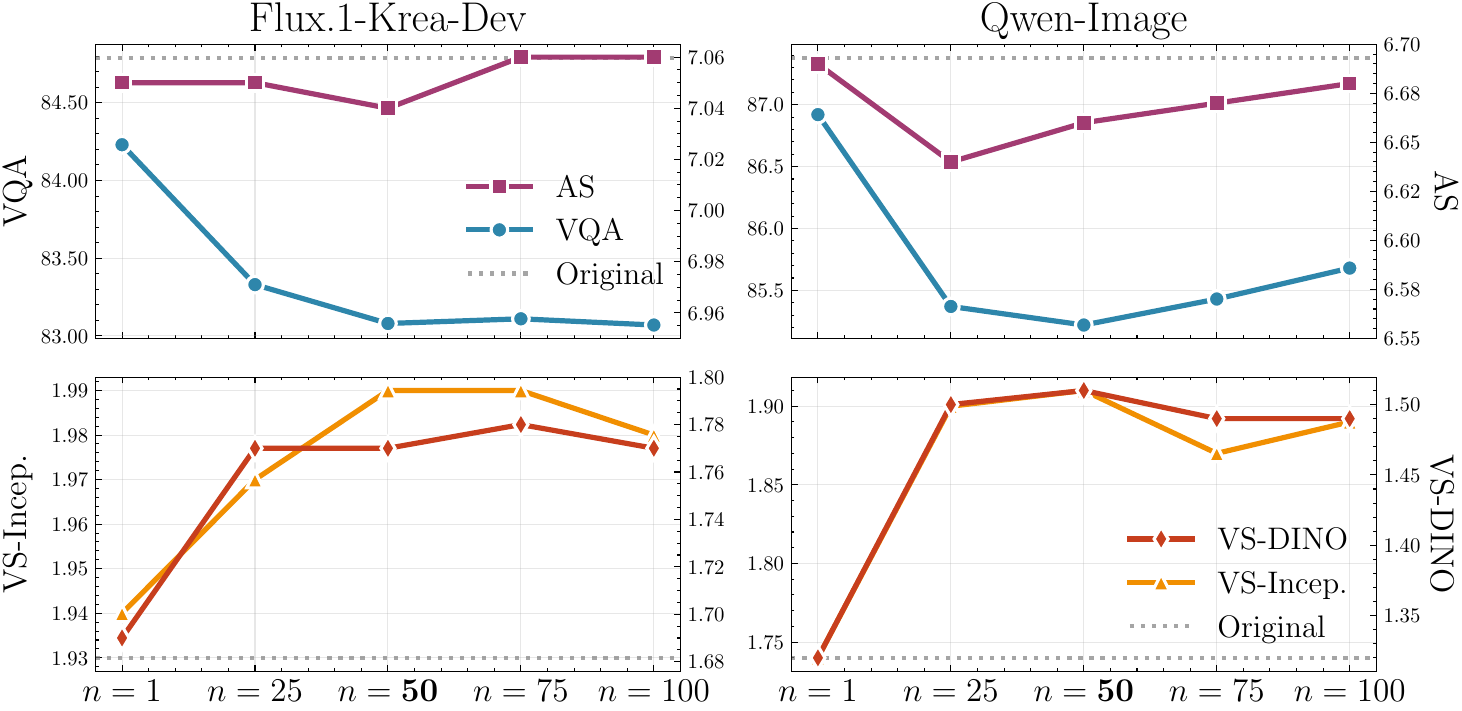}
        \vspace{-8.5pt}
        \captionof{figure}{\textbf{Comparison across different $n$.} The default setting is highlighted in \textbf{bold}. Dashed line: w/o. \abbrev.}
        \label{fig:abla_mode}
    \end{minipage}
    \hfill
    \begin{minipage}[t]{0.49\linewidth}
        \centering
        \vspace{-85pt}
        \includegraphics[width=\linewidth]{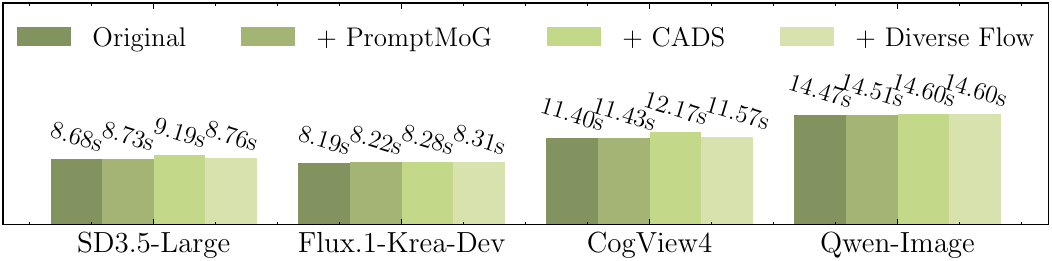}
        \includegraphics[width=\linewidth]{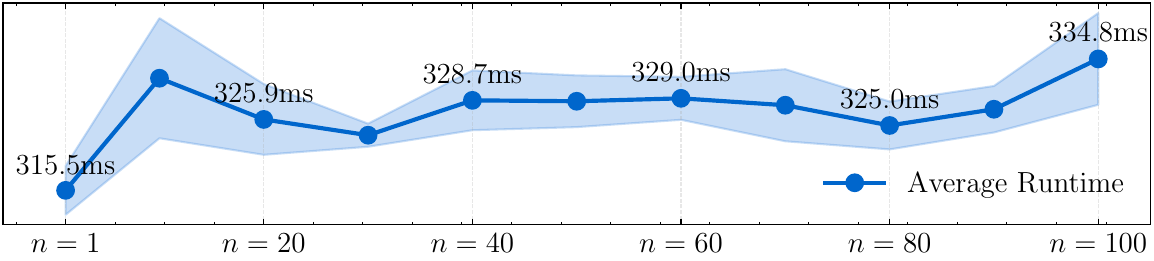}
        \vspace{-10pt}
        \captionof{figure}{\textbf{Runtime analysis.} \textbf{(Top):} Runtime across approaches. \textbf{(Bottom):} Runtime under different modes ($n$).}
        \label{fig:runtime}
    \end{minipage}

    \vspace{-5pt}
\end{figure}

\subsection{Ablation Study}\label{sec:abla}

\paragraph{\textbf{Settings.}} For efficient evaluation, we sample eight prompts per theme in LPD-Bench, yielding \textbf{200} prompts for ablation. We use Flux.1-Krea-Dev and Qwen-Image as representative encoder-based and decoder-based models for text embeddings. Results are reported as averages over VQA, AS, VS-Incep. and VS-DINO.
\vspace{-10pt}
\paragraph{\textbf{Effect of $\gamma_{\text{sim}}$.}} $\gamma_{\text{sim}}$ defines the feasible region on the high-dimensional sphere. The results of different values are in~\cref{fig:abla_gamma}. A higher $\gamma_{\text{sim}}$ corresponds to a smaller $\gamma_{\text{euc}}$ as shown in~\cref{eq:sim_to_euc}, thereby reducing the feasible region. Consequently, reformulated prompts remain closer to the original prompt, leading to higher semantic consistency as reflected by increasing VQA scores. For aesthetics, a smaller sampling space prevents deviation into irrelevant regions, resulting in stable or improved aesthetic quality, particularly for Qwen-Image. However, the restricted sampling region limits diversity, yielding lower VS scores.
\vspace{-10pt}
\paragraph{\textbf{Effect of $\sigma$.}} $\sigma$ controls the sampling range of each Gaussian component. The results of different values are in~\cref{fig:abla_sigma}. A larger $\sigma$ expands the sampling space, enhancing diversity as evidenced by higher VS-Incep. and VS-DINO. However, an excessively large $\sigma$ increases the chance of sampling embeddings that deviate from the semantic center, causing slight semantic drift and lower VQA scores. Since the overall distance to the center remains constrained, the aesthetics score remains nearly stable.
\vspace{-10pt}
\paragraph{\textbf{Effect of $n$.}} $n$ determines the number of reformulated prompts used to form the MoG. The results of different values are in~\cref{fig:abla_mode}. As predicted by our theoretical analysis, increasing $n$ enhances diversity from $n=1$ to $n=50$. When $n$ becomes large (\eg, $n=75$ or $n=100$), different Gaussian components begin to overlap, violating the disjointness and removing the guarantee of entropy increase. Notably, variations in $n$ have minimal influence on semantic fidelity and aesthetics, as both are effectively constrained by $\gamma_{\text{sim}}$ and $\sigma$.

\subsection{Runtime Analysis}

\paragraph{\textbf{Between Methods.}} We compare the runtime across different approaches, as shown in~\cref{fig:runtime}, top. Each method is executed 5 times, and the average runtime is reported. Overall, different approaches adopt efficient designs. However, our method is slightly faster because prompt embedding sampling is performed only once at the beginning, whereas other approaches require perturbation or optimization at every denoising step.
\vspace{-10pt}
\paragraph{\textbf{Across Modes.}} A potential concern is that runtime may increase as the number of reformulated sets (\ie, modes) grows. As illustrated in~\cref{fig:runtime}, bottom, this is not the case. By vectorizing the prompt embedding sampling, our method maintains an almost constant runtime with only a marginal increase as the number of modes grows, demonstrating strong scalability and efficiency.

\section{Conclusion and Future Work}

In this work, we address the challenge of reduced diversity in long-prompt image generation. With our proposed LPD-Bench, we demonstrate that large-scale T2I models consistently exhibit a decline in diversity as prompt length increases. To mitigate this issue, we develop a theoretical framework that connects prompt reformulation to entropy and introduce a plug-and-play method, \abbrev, that samples prompt embeddings from a Mixture-of-Gaussians to enhance diversity while maintaining semantic fidelity. Extensive experiments on four large-scale models, including SD3.5 Large, Flux.1-Krea-Dev, CogView4, and Qwen-Image, as well as on LDM and SD2.1, validate the effectiveness and generalizability of \abbrev. We further demonstrate that \abbrev\ applies effectively to large-scale autoregressive models, including Show-o2 and BLIP3o-NEXT. For future work, we plan to extend our formulation to other modalities, \eg, images and audio, to enable cross-modal diversification and perform space truncation to retain only the most informative subspace.

\section*{Acknowledgment}

This work is partially supported by the National Science and Technology Council, Taiwan, under Grant: NSTC-115-2628-E-A49 -019-MY4 and NSTC-115-2640-B-005-001.

\bibliography{main}
\clearpage
\begin{appendices}
    \appendix

{
\centering
\LARGE
\textbf{\textcolor{bmv@sectioncolor}{Appendix}} \\
}

\section{Reproducibility Statement}

We provide our code in the supplementary materials. To ensure similar results across different hardware configurations, we generate random noise on the CPU so that runs with the same seed produce \textit{nearly} identical outputs\footnote{GPU operations may still be non-deterministic.}.

\section{LPD-Bench Creation}\label{sec:dataset_details}
\subsection{Data Construction and Filtering}

\begin{figure}[h]
    \centering
    \includegraphics[width=\linewidth]{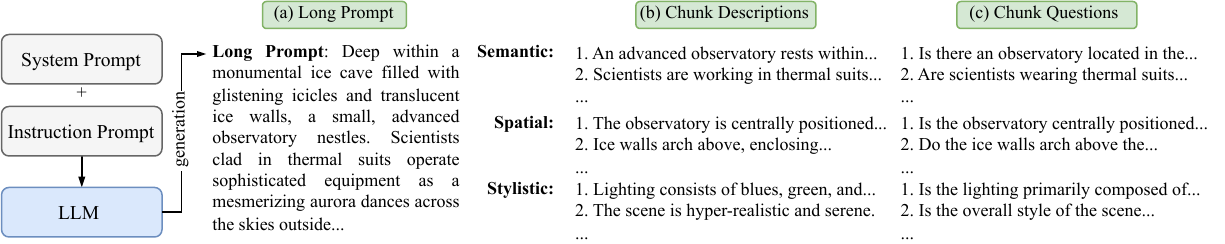}
    \caption{\textbf{Example of a generated long-prompt.} 
    Each sample includes a long prompt along with chunk-level descriptions and yes/no questions (answers are always yes) covering the \textit{semantic}, \textit{spatial}, and \textit{stylistic} aspects for evaluation.}
    \label{fig:data_collection}
\end{figure}

We construct LPD-Bench through three stages: \textbf{(1)} deciding the themes, \textbf{(2)} generating a large pool of candidate prompts, and \textbf{(3)} filtering the pool to retain a diverse subset with minimal redundancy. An example of the generated output is shown in~\cref{fig:data_collection}.

\paragraph{\textbf{Prompt Themes.}} We adopt the photography taxonomy from Adobe LightRoom~\cite{adobe_photography_types}, which initially defines 28 categories. We consolidate these into 25 representative themes by removing overlapping categories (\eg, merging ``Portrait'' and ``Headshot'' into a single category). The final themes are listed in~\cref{tab:topics}.

\begin{table}[h]
\centering
\begin{tabular}{ccccc}
\toprule
Landscape & Seascape & Wildlife & Macro & Astrophotography \\
\midrule
Weather & Botanical & Architecture & Urban & Real Estate \\
\midrule
Aerial & Portrait & Fashion & Sports & Candid \\
\midrule
Event & Food & Product & Still Life & Automotive \\
\midrule
Fine Art & Black and White & Abstract & Surreal & Long Exposure \\
\bottomrule
\end{tabular}
\caption{\textbf{Categories of LPD-Bench.}}
\label{tab:topics}
\end{table}
\vspace{-8pt}
\paragraph{\textbf{Pool Generation.}} Our system and instruction prompts, following recommended practice from~\cite{oppenlaender2024taxonomy}, are shown in the colored boxes below to generate an initial pool of \textbf{60} samples for each topic. Each sample includes a long prompt together with its corresponding chunked descriptions and yes/no questions across the \textit{semantic}, \textit{spatial}, and \textit{stylistic} aspects.
% \vspace{-8pt}
\paragraph{\textbf{Prompt Filtering.}} To reduce redundancy and ensure diversity, we perform pairwise similarity-based filtering to select a subset of \textbf{40} prompts that are maximally distinct for each theme. Formally, let $\mathcal{C} = \{c_i\}_{i=1}^{N}$ denote the complete set of generated prompts, where each prompt $c_i$ contains three descriptive components: semantic, spatial, and stylistic. We embed each component using a pretrained sentence encoder $f_{\theta}(\cdot)$ to obtain:
\begin{equation}
\mathbf{s}_i = f_{\theta}(c_i^{\text{semantic}}), \quad
\mathbf{r}_i = f_{\theta}(c_i^{\text{spatial}}), \quad
\mathbf{t}_i = f_{\theta}(c_i^{\text{stylistic}}).
\end{equation}
The final representation of a prompt is the concatenation of the normalized embeddings:
\begin{equation}
\mathbf{e}_i = \operatorname{concat}\big[
\operatorname{norm}(\mathbf{s}_i),\,
\operatorname{norm}(\mathbf{r}_i),\,
\operatorname{norm}(\mathbf{t}_i)
\big] \in \mathbb{R}^{(3 \times d)}.
\end{equation}
We then compute the cosine similarity matrix $S_{ij}$ and the average similarity $\bar{s}_i$ for each prompt:
\begin{equation}
S_{ij} = \frac{\mathbf{e}_i^{\top}\mathbf{e}_j}{\|\mathbf{e}_i\|_2 \, \|\mathbf{e}_j\|_2}, 
\quad
\bar{s}_i = \frac{1}{N-1} \sum_{j \neq i} S_{ij}.
\end{equation}
A smaller $\bar{s}_i$ indicates that a prompt is less correlated with others. The final subset $\mathcal{C}_{\text{sel}}$ consists of $K=40$ prompts corresponding to the lowest $\bar{s}_i$ values:
\begin{equation}
\mathcal{C}_{\text{sel}} = \{\, c_i \mid i \in \operatorname{argsort}(\bar{s})_{1:K} \,\}.
\end{equation}

\begin{prompt}{System Prompt for Long Prompt Generation}
    \small
    You are an expert visual prompt engineer and researcher specializing in generating long, compositionally rich prompts for text-to-image tasks.
\end{prompt}

\begin{prompt}{Instruction Prompt for Long Prompt Generation}
\small
\textbf{Your task is to:}
\begin{enumerate}
    \item Write a long, detailed, and coherent text-to-image prompt based on the given theme
    \item Ensure the prompt is not similar to any of the existing prompts
    \item Decompose the scene into \textbf{semantic}, \textbf{spatial}, and \textbf{stylistic} aspects
    \item For each aspect, provide \textbf{chunked descriptions} and \textbf{diverse QA pairs} for evaluation
\end{enumerate}

\vspace{0.3cm}
\noindent\textbf{1. Theme}

\noindent Theme: \texttt{\{theme\}}\\
Description: \texttt{\{description\}}\\
Existing prompts: \texttt{\{existing\_prompts\}}

\vspace{0.3cm}
\noindent\textbf{2. Prompt Creation Rules}

\noindent\textit{Overall Structure \& Length}
\begin{itemize}[leftmargin=*,noitemsep,topsep=2pt]
    \item Target length: 250--450 words
    \item Format: Coherent narrative paragraph(s), NOT keyword lists
    \item Tone: Third-person, present-tense, cinematic description
    \item Coherence: Maintain internal consistency throughout
\end{itemize}

\noindent\textit{Required Components}

\noindent Your prompt MUST include these six elements in natural language:

\noindent\textbf{(1) Subject Terms:} Main subject(s) and their characteristics, environment and setting details, subject relationships and activities.

\noindent\textbf{(2) Composition \& Spatial Arrangement:} Camera perspective, framing, depth (foreground, midground, background), scale relationships.

\noindent\textbf{(3) Lighting \& Atmosphere:} Light source, light quality, atmospheric conditions, shadows, and highlights.

\noindent\textbf{(4) Color Palette \& Mood:} Dominant colors, color relationships, color temperature, emotional tone.

\noindent\textbf{(5) Style Modifiers:} Artistic style, medium characteristics, technical approach.

\noindent\textbf{(6) Quality Boosters:} Technical excellence markers, professional indicators, recognition markers.

\vspace{0.3cm}
\noindent\textbf{3. Description and QA Generation Rules}

\noindent For each aspect---\textbf{semantic}, \textbf{spatial}, and \textbf{stylistic}:

\noindent\textit{Description Chunks:} Provide 2--4 short declarative statements (10--25 words each). Each chunk describes ONE observable visual fact.

\noindent\textit{Yes/No Questions:} Provide 2--4 questions per aspect. All correct answers must be ``Yes''. Questions must be unambiguous, verifiable, aspect-aligned, affirmative, and specific.
\end{prompt}

\subsection{Data Statistics}\label{sec:data_statistics}

To analyze the linguistic composition of the benchmark datasets, we compute several statistics based on part-of-speech and phrase-level analyses. 
For each prompt $c$, we first process its text using a pretrained spaCy model to obtain a tokenized sequence $\{w_i\}_{i=1}^{N_c}$, excluding punctuation and stop words. 
Each token or phrase is then categorized into one of two aspects, \emph{spatial} or \emph{stylistic}, as illustrated in~\cref{fig:dataset_lengths}. 
Multi-word phrases are detected using spaCy’s \texttt{PhraseMatcher}\footnote{\url{https://github.com/explosion/spaCy}} with predefined spatial and stylistic phrase lists generated by \texttt{GPT-4o}, while single tokens are tagged according to their membership in the corresponding lexical sets.

Let $n_{\text{spa}}(c)$ and $n_{\text{sty}}(c)$ denote the number of spatial and stylistic tokens (or phrase spans) identified in prompt $c$. 
We define their normalized densities as:
\begin{equation}
d_{\text{spa}}(c) = \frac{n_{\text{spa}}(c)}{N_c}, 
\qquad 
d_{\text{sty}}(c) = \frac{n_{\text{sty}}(c)}{N_c},
\end{equation}
where $N_c$ is the total number of valid tokens in the prompt.

A prompt is considered to \emph{cover} an aspect if it satisfies either a minimum density or a minimum count criterion:
\begin{equation}
\text{cover}_{\text{spa}}(c) = \bm{1}_\mathrm{\big\{d_{\text{spa}}(c) \ge \tau_{\text{spa}} \;\lor\; n_{\text{spa}}(c) \ge k_{\text{spa}}\big\}},
\end{equation}
where $\bm{1}_\mathrm{\{\cdot\}}$ is the indicator function and analogously for the stylistic category.  
We set $\tau_{\text{spa}} = \tau_{\text{sty}} = 0.05$ and $k_{\text{spa}} = k_{\text{sty}} = 5$ in all experiments.  
The overall dataset-level \emph{spatial} and \emph{stylistic coverage} are computed as the proportions of prompts that satisfy each condition.

To quantify the balance between the two aspects within a prompt, we compute a normalized entropy score:
\begin{equation}
H(c) = -\sum_{a \in \{\text{spa}, \text{sty}\}} 
\tilde{d}_a(c)\,\log \tilde{d}_a(c),
\end{equation}
where $\tilde{d}_a(c) = d_a(c) / (d_{\text{spa}}(c) + d_{\text{sty}}(c))$.  
The balance index $H(c) \in [0, 1]$ equals $1$ when spatial and stylistic terms are evenly distributed and approaches $0$ when dominated by a single category.  
We report the dataset-level average $\bar{H} = \frac{1}{|\mathcal{C}|}\sum_{c \in \mathcal{C}} H(c)$ as the \emph{balance score}.  
Additionally, we measure the \emph{average prompt length} using the number of characters.

\subsection{Overall Score}\label{sec:overall_score}

To quantify the general quality-diversity trade-off, we define:
\begin{equation}
    \text{Overall Score} =
    (-\Delta\text{VQA} + - \Delta \text{AS})/
    (\Delta \text{VS-Incep.} + \Delta \text{VS-DINO}),
\end{equation}
where $\Delta(\cdot)$ denotes the change of each method relative to the original model. The numerator captures semantic degradation in VQA and AS, with VQA normalized to $[0,1]$, while the denominator aggregates diversity gains measured by two Vendi scores. A lower score indicates a better tradeoff, corresponding to larger diversity gains with smaller semantic drops.
\section{Proof of Statement}

\subsection{Entropy Comparison for Long and Short Prompts}\label{sec:proof_long_short}

Let $\mathbf{X}$ be a random variable representing generated samples, and let $c_{s}$ and $c_{l}$ denote the short prompt and the long prompt, respectively. Assume that the long prompt contains all the information of the short prompt plus additional details $u$, \ie:
\begin{equation}
    c_{l} = (c_{s}, u).
\end{equation}
We show that the conditional entropy under the long prompt is no larger than that under the short prompt:
\begin{equation}
    H(\mathbf{X}\mid c_{l}) \le H(\mathbf{X}\mid c_{s}).
\end{equation}

\begin{proof}
By the definition of conditional mutual information, we have:
\begin{equation}
    I(\mathbf{X}; u\mid c_{s})
      = H(\mathbf{X}\mid c_{s}) - H(\mathbf{X}\mid c_{s}, u).
\end{equation}
After rearranging, we have:
\begin{equation}
    H(\mathbf{X}\mid c_{s}, u)
      = H(\mathbf{X}\mid c_{s}) - I(\mathbf{X}; u\mid c_{s}).
\end{equation}
Since mutual information is always nonnegative, $I(\mathbf{X}; u\mid c_{s}) \ge 0$, it follows that:
\begin{equation}
    H(\mathbf{X}\mid c_{l})
      = H(\mathbf{X}\mid c_{s}, u)
      \le H(\mathbf{X}\mid c_{s}).
\end{equation}
Therefore, conditioning on a longer (more informative) prompt can only reduce or leave unchanged the uncertainty in $\mathbf{X}$.
\end{proof}

\subsection{Entropy with Reformulated Prompt}\label{sec:proof_prompt_reformulate}

Let $\mathbf{X}$ be a random variable representing the generated samples, and let $c$ and $c'$ denote the target prompt and a reformulated prompt, respectively. We define $\mathbf{C}_{\text{ref}}$ as the random variable corresponding to the set of possible reformulated prompts $\mathcal{C}_{\text{ref}}$, whose distribution is given by $p(c'\mid c)$. That is, $\mathbf{C}_{\text{ref}}$ models the stochastic process of drawing a reformulated prompt $c'$ conditioned on the target prompt $c$. Additionally, $I(\mathbf{X}; \mathbf{C}_{\text{ref}}\mid c)$ is the conditional mutual information between the generated sample and the reformulated prompt given the original prompt $c$.

We show that the conditional entropy under the target prompt can be expressed as a weighted combination of the conditional entropies under the reformulated prompts:
\begin{equation}
    H_n(\mathbf{X}\mid c)
    = \sum_{c'\in\mathcal{C}_{\text{ref}}} p(c'\mid c)\,H(\mathbf{X}\mid c')
    + I(\mathbf{X};\mathbf{C}_{\text{ref}}\mid c),
\end{equation}
where $H_n$ denotes the entropy computed when the cardinality $|\mathcal{C}_{\text{ref}}| = n$.

\begin{proof}
From the chain rule of conditional entropy, we have:
\begin{equation}
    H(\mathbf{X},\mathbf{C}_{\text{ref}}\mid c)
    = H(\mathbf{X}\mid \mathbf{C}_{\text{ref}}, c)
    + H(\mathbf{C}_{\text{ref}}\mid c)
    = H(\mathbf{C}_{\text{ref}}\mid \mathbf{X}, c)
    + H(\mathbf{X}\mid c).
\end{equation}
Rearranging the last two expressions yields:
\begin{equation}
    H(\mathbf{X}\mid c)
    = H(\mathbf{X}\mid \mathbf{C}_{\text{ref}}, c)
    + H(\mathbf{C}_{\text{ref}}\mid c)
    - H(\mathbf{C}_{\text{ref}}\mid \mathbf{X}, c).
\end{equation}
Recognizing that the last two terms form the conditional mutual information, we obtain:
\begin{equation}
    H(\mathbf{X}\mid c)
    = H(\mathbf{X}\mid \mathbf{C}_{\text{ref}}, c)
    + I(\mathbf{X};\mathbf{C}_{\text{ref}}\mid c),
\end{equation}
where:
\begin{equation}
    I(\mathbf{X};\mathbf{C}_{\text{ref}}\mid c)
    = H(\mathbf{C}_{\text{ref}}\mid c)
    - H(\mathbf{C}_{\text{ref}}\mid \mathbf{X}, c).
\end{equation}
Finally, by expanding $H(\mathbf{X}\mid \mathbf{C}_{\text{ref}}, c)$ over the discrete set $\mathcal{C}_{\text{ref}}$, we have:
\begin{equation}
    H(\mathbf{X}\mid c)
    = \underbrace{
        \sum_{c'\in\mathcal{C}_{\text{ref}}}
        p(c'\mid c)\,H(\mathbf{X}\mid c')
    }_{H(\mathbf{X}\mid \mathbf{C}_{\text{ref}}, c)}
    + I(\mathbf{X};\mathbf{C}_{\text{ref}}\mid c),
\end{equation}
which completes the proof.
\end{proof}

\subsection{Monotonic Increasing Property}\label{sec:proof_increasing}

Let $\mathbf{X}$ be a random variable representing the generated samples $\mathbf{x}$, and let $c$ and $c'$ denote the target prompt and a reformulated prompt from a set $\mathcal{C}^{(n)}_{\text{ref}} \triangleq \{c'_1, \cdots, c'_n \}$, respectively.

\paragraph{\textbf{Assumptions:}}
\begin{enumerate}[leftmargin=1cm]
    \item[(A1)] (\emph{Equal component entropies}) $H(\mathbf{X}\mid c'_i)=h$ for all $i$.
    \item[(A2)] (\emph{Disjoint supports}) The conditional densities $p(\mathbf{X}\mid c'_i)$ have pairwise disjoint supports $S_i$, \ie, $S_i\cap S_j=\varnothing$ for $i\neq j$.
    \item[(A3)] (\emph{Uniform weights}) $p(c'_i\mid c)=1/n$ for all $i$.
\end{enumerate}

With the assumptions above, we have the monotonic increasing property:
\begin{equation}
    H_{n+1}(\mathbf{X}\mid c) - H_n{(\mathbf{X}\mid c)} = \log(n+1) - \log(n) > 0.
\end{equation}

\begin{proof}

By (A3), for any $\mathbf{x}\in S_i$ we have $p(\mathbf{X}\mid c) = \tfrac{1}{n} p(\mathbf{X}\mid c'_i)$, and $p(\mathbf{X}\mid c'_j)=0$ for $j\neq i$.
Therefore:
\begin{align}
    H_n(\mathbf{X}\mid c)
    &= -\int p(\mathbf{X}\mid c)\,\log p(\mathbf{X}\mid c)\,d\mathbf{x} \\
    \overset{\text{By (A2)}}&{=} -\sum_{i=1}^n \int_{S_i} \frac{1}{n} p(\mathbf{X}\mid c'_i)\,
       \log\!\Big(\tfrac{1}{n}p(\mathbf{X}\mid c'_i)\Big)\,d\mathbf{x} \\
    &= -\frac{1}{n}\sum_{i=1}^n \int_{S_i} p(\mathbf{X}\mid c'_i)\,\log p(\mathbf{X}\mid c'_i)\,d\mathbf{x}
       \;+\; \frac{1}{n}\sum_{i=1}^n \int_{S_i} p(\mathbf{X}\mid c'_i)\,\log n \, d\mathbf{x} \\
    &= \frac{1}{n}\sum_{i=1}^n H(\mathbf{X}\mid c'_i) \;+\; \log n.
\end{align}
By (A1), $\frac{1}{n}\sum_{i=1}^n H(\mathbf{X}\mid c'_i)=h$, hence $H_n(\mathbf{X}\mid c)=h+\log n$. The stated difference $H_{n+1}-H_n=\log(n+1)-\log n>0$ follows immediately, proving strict monotonicity.
\end{proof}

\subsection{Regular Simplex}\label{sec:proof_simplex}

Given a base embedding $\boldsymbol{e}_c\in\mathbb{R}^d$, a radius $\gamma>0$, and a target number of variants $n\ge2$, construct centers $\{\mathbf{\mu}_i\}_{i=1}^n\subset\mathbb{R}^d$ such that:
\begin{equation}
    \|\mathbf{\mu}_i-\boldsymbol{e}_0\|_2=\gamma
    \quad\text{and}\quad
    \mathbf{\mu}_i^\top \mathbf{\mu}_j = -\gamma^2\! / (n-1) \ (i\neq j),
\end{equation}
\ie, all points lie on the hypersphere of radius $\gamma$ around $\boldsymbol{e}_c$ and are maximally/equally separated.

\paragraph{\textbf{Constraint:}}
Let $\{ \boldsymbol{e}_{c_i'} \}_{i=1}^n$ satisfy:
\begin{equation}
    \|\boldsymbol{e}_{c'_i}\|_2=1,\quad
    \boldsymbol{e}_{c'_i}^\top \boldsymbol{e}_{c'_j} =
    \begin{cases}
        1,& i=j,\\[2pt]
        -\frac{1}{n-1},& i\neq j.
    \end{cases}
\end{equation}
These are the vertices of a regular simplex on the unit sphere $S^{n-2}$.
They achieve equal pairwise angles $\arccos\!\big(-\tfrac{1}{n-1}\big)$ and equal pairwise distances.

Let $I_n$ be the $n\times n$ identity and $\mathbf{1}_n$ the all-ones vector. Define the $n\times n$ centering matrix $H=I_n-\tfrac{1}{n}\mathbf{1}_n\mathbf{1}_n^\top$ and set:
\begin{equation}
    V = H \in \mathbb{R}^{n\times n}.
\end{equation}
Let the thin SVD be $V=U\Sigma W^\top$ and take the first $n-1$ right singular vectors $W_{1:(n-1)}$.
Then the $n$ rows of
\begin{equation}
    \tilde U \;=\; W_{1:(n-1)} \in \mathbb{R}^{n\times (n-1)},
\end{equation}
form a centered and orthonormal configuration.
Normalize each row to unit norm to obtain $\mathbf{u}_1,\ldots,\mathbf{u}_n$ with the simplex inner products above. 
For any $d\ge n-1$, pad each $\mathbf{u}_i$ with zeros to $\hat{\mathbf{u}}_i\in\mathbb{R}^d$, apply an arbitrary orthogonal transform $R\in\mathrm{O}(d)$ (\eg, a rotation matrix), and set:
\begin{equation}
    \boldsymbol{e}_{c'_i} \;=\; \boldsymbol{e}_c + \gamma\, R\, \hat{\mathbf{u}}_i
    \quad\text{for } i=1,\dots,n.
\end{equation}
Then $\|\boldsymbol{e}_{c'_i}-\boldsymbol{e}_c\|_2=\gamma$ and
$\boldsymbol{e}_{c'_i}^\top \mathbf{c}_{c'_j}=-\gamma^2/(n-1)$ for $i\ne j$.

\section{More Analysis between Prompt Length and Diversity}\label{sec:prompt_length_diversity}

\begin{figure}[t]
    \centering
    \includegraphics[width=\linewidth]{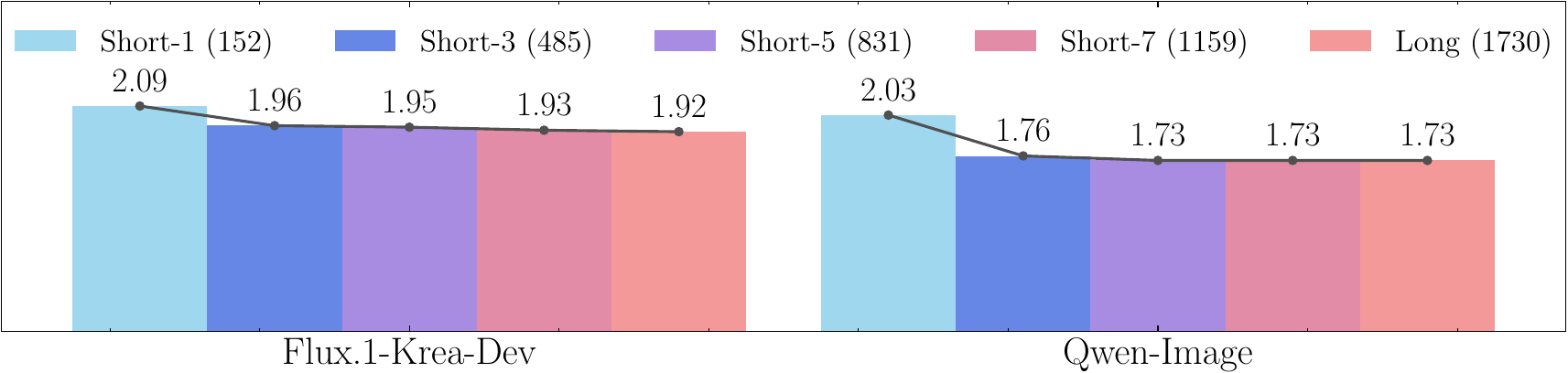}
    \caption{\textbf{Finer diversity analysis with additional prompt-length intervals.}}
    \label{fig:deeper_prompt_diversity_analysis}
\end{figure}

We observe in~\cref{sec:method_diverse,fig:diversity_comparison} that diversity degrades as prompt length increases. To analyze this trend more finely, we further divide prompts into additional length intervals by using the first five and first seven sentences from each prompt in LPD-Bench, following the setting in~\cref{fig:diversity_comparison}. As shown in~\cref{fig:deeper_prompt_diversity_analysis}, diversity decreases in a log-like manner as prompt length increases, \ie it drops rapidly at first and then gradually saturates.

Intuitively, longer prompts impose more simultaneous constraints on the generated image. Each additional token narrows the set of images that satisfy the prompt, which effectively shrinks the feasible solution space. As a result, the conditional distribution $p(\mathbf{X}\mid c)$ becomes increasingly concentrated, and the randomness available during sampling decreases. When the prompt becomes sufficiently long, newly added tokens often provide partially redundant or overlapping constraints, so their marginal effect becomes smaller and the diversity curve gradually saturates.

To formalize this intuition, let $\ell$ denote the prompt length and let $s(\ell)$ denote the constraint strength induced by a prompt of length $\ell$. Intuitively, $s(\ell)$ measures how strongly the prompt restricts the set of valid generations. We assume that each additional token still increases the total constraint strength, but its marginal contribution decreases as the prompt becomes longer. This yields:
\begin{equation}
    \frac{d s(\ell)}{d \ell} = \frac{r}{\ell + \ell_0},
\end{equation}
where $r > 0$ controls the overall rate at which new tokens add constraints, and $\ell_0 > 0$ is a small offset that avoids singular behavior for short prompts and can be interpreted as a base prompt scale. Integrating gives:
\begin{equation}
    s(\ell) = s_0 + r \log\!\left(1 + \frac{\ell}{\ell_0}\right),
\end{equation}
where $s_0$ denotes the base constraint strength at the reference prompt scale. Intuitively, $s_0$ captures the amount of conditioning already present before prompt length begins to substantially increase.

We then relate this constraint strength to the conditional entropy of valid generations. Specifically, we assume constraint strength reduces entropy approximately linearly:
\begin{equation}
    H(\mathbf{X} \mid c_\ell)
    =
    H_0 - m \cdot s(\ell),
\end{equation}
where $H(\mathbf{X} \mid c_\ell)$ denotes the conditional entropy of generated images under a prompt of length $\ell$, $H_0$ denotes the unconstrained or weakly constrained entropy level, and $m > 0$ measures how strongly prompt constraints reduce entropy. Intuitively, a larger $H_0$ means the model has more freedom when conditioning is weak, while a larger $m$ means the model becomes concentrated more quickly as constraints accumulate. Substituting $s(\ell)$ gives:
\begin{equation}
    H(\mathbf{X} \mid c_\ell)
    =
    H_0 - m\cdot s_0 - m \cdot r \log\!\left(1 + \frac{\ell}{\ell_0}\right).
\end{equation}

Since generative diversity is positively related to conditional entropy, we further obtain the diversity score by:
\begin{equation}
    D(\ell) = D_0 - \alpha \log\!\left(1 + \frac{\ell}{\ell_0}\right),
\end{equation}
where $D(\ell)$ denotes the diversity at prompt length $\ell$, $D_0$ denotes the diversity level for short prompts, and $\alpha > 0$ controls how quickly diversity decreases as prompt length increases. This expression captures the empirical trend in~\cref{fig:deeper_prompt_diversity_analysis} with log-scale: prompt length initially reduces diversity rapidly, while the effect gradually weakens as newly added tokens become more redundant. A rigorous characterization would require a principled definition of prompt length, token redundancy, and effective constraint strength. We leave this direction for future work.

\section{Empirical Validation of Theory}

\subsection{Disjointness of Gaussian Components}\label{sec:disjointness}

To connect the theoretical assumptions with the practical construction of our MoG, we empirically evaluate the overlap among Gaussian components. Since the disjointness assumption requires samples from different components to occupy nearly nonoverlapping regions, we first measure the nearest center cross component rate:
\begin{equation}
p\!\left[
\arg\min_j \|\mathbf{x}-\boldsymbol{\mu}_j\|_2 \neq i
\;\middle|\;
\mathbf{x}\sim \mathcal{N}(\boldsymbol{\mu}_i,\sigma^2 \mathbf{I})
\right].
\end{equation}
This metric quantifies the probability that a sample drawn from the $i$th Gaussian component is closer to another component center than to its own center. We further compute the Bhattacharyya coefficient between each pair of isotropic Gaussian components:
\begin{equation}
\mathrm{BC}_{ij}=
\exp\!\left(
-\frac{\|\boldsymbol{\mu}_i-\boldsymbol{\mu}_j\|_2^2}{8\sigma^2}
\right),
\end{equation}
which directly measures their distributional overlap.

Under the default setting with $n=50$ components and $10{,}000$ samples drawn from each component, the cross component rate is $0.00\%$ for both $D=1024$ and $D=4096$. The maximum Bhattacharyya coefficient is also negligible, with
$\log_{10}\max_{i\neq j}\mathrm{BC}_{ij}\approx -1815$ for $D=1024$ and
$\log_{10}\max_{i\neq j}\mathrm{BC}_{ij}\approx -7260$ for $D=4096$.
These results indicate that the Gaussian components are effectively disjoint in the embedding space, thereby empirically supporting the disjointness assumption used in our entropy analysis. Moreover, our construction places all component centers at the same distance from the original prompt embedding. Consequently, each reformulated conditioning remains subject to the same semantic proximity constraint, which supports the assumption that the corresponding conditional entropies remain comparable.

\subsection{Identical Norm Assumption}\label{sec:norm}

In~\cref{eq:sim_to_euc}, $\gamma_{\mathrm{sim}}$ specifies the desired cosine similarity to the original prompt embedding, while $\boldsymbol{e}_{c'_i}$ is constructed only after the radius is determined. Therefore, we adopt the equal norm approximation to eliminate the unknown norm term and convert the cosine constraint into the Euclidean radius $\gamma_{\mathrm{euc}}$. We empirically examine this approximation by comparing the norm of sampled \abbrev\ embeddings with that of the original prompt embedding. Under the default setting, the mean norm ratio is $1.14$ for T5, indicating only a modest deviation from the assumption. This result suggests that the sampled embeddings remain close in magnitude to the original embedding, supporting the validity of the equal norm approximation used throughout our analysis.
\section{Setup and Additional Comparison}

\subsection{Hyperparameter Setup}\label{sec:hyper}

We summarize the hyperparameters used in our experiments in~\cref{tab:hyper}. For CogView4 and Qwen-Image, which employ transformer decoders to produce text embeddings, we adopt a larger $\gamma$. In contrast to encoder-based models, decoder-based embeddings exhibit weaker semantic locality, and an overly small $\gamma$ leads to unstable or degraded generations. This phenomenon is further analyzed in~\cref{sec:abla}. For SD2.1, we follow SPARKE~\cite{jalali2025sparke} and adopt DPMSolver~\cite{lu2025dpm} for denoising.

{
\setlength{\tabcolsep}{7pt}
\begin{table}[h]
\centering
\begin{tabular}{lccccc}
\toprule
Model & $\gamma$ & $\sigma$ & $n$ & Resolution & Inference Step \\
\midrule
LDM~\cite{rombach2022high} & 0.60 & 0.20 & 50 & $256\times256$  & 50 \\
SD2.1~\cite{rombach2022high} & 0.60 & 0.20 & 50 & $1024\times1024$ & 50 \\
SD3.5-Large~\cite{peebles2023scalable} & 0.70 & 0.25 & 50 & $1024\times1024$ & 28 \\
Flux.1-Krea-Dev~\cite{flux1kreadev2025} & 0.60 & 0.25 & 50 & $1024\times1024$  & 28 \\
CogView4~\cite{zheng2024cogview} & 0.95 & 0.20 & 50 & $1024\times1024$ & 28 \\
Qwen-Image~\cite{wu2025qwen} & 0.85 & 0.25 & 50 & $1024\times1024$ & 28 \\
Show-o2~\cite{xie2025showo} & 0.60 & 0.30 & 50 & $1024\times1024$ & - \\
BLIP3o-NEXT~\cite{chen2025blip3o} & 0.60 & 0.30 & 50 & $1024\times1024$ & - \\
\bottomrule
\end{tabular}
\caption{\textbf{Hyperparameter settings for different models.}}
\label{tab:hyper}
\end{table}
}

\subsection{Prompt Chunking and Expansion}\label{sec:prompt_chunking}

\paragraph{\textbf{Chunking.}} We adopt prompt chunking based on our earlier observation that generation diversity decreases as prompt length increases (see~\cref{fig:diversity_comparison}). To address this, we split each prompt into several overlapping chunks and average their embeddings. This strategy effectively simulates using shorter textual segments to represent a long prompt.  Formally, given a prompt $c$ consisting of $k$ sentences $[c_{s_1}, c_{s_2}, \cdots, c_{s_k}]$, we define a window size $w$ with a stride of one to group sentences, forming $k - w$ overlapping chunks. We then compute the final embedding $\boldsymbol{e}_{c'}$ by averaging the embeddings of all chunks:
\begin{equation}
    \boldsymbol{e_{c'}} = \frac{1}{k-w}\sum_{i=1}^{k-w}E(\hat{c}_i), \quad \text{with} \quad \hat{c}_i \triangleq [c_{s_i}, \cdots, c_{s_{i+w}}],
\end{equation}
where $E$ denotes a text encoder.
\vspace{-8pt}
\paragraph{\textbf{Expansion}.} As PE does not release its official weights and training dataset, we instead employ an LLM to rephrase each original prompt into multiple variations and use these different prompts to generate outputs.

\subsection{Additional VQA Model}\label{sec:add_vqa}

{
\renewcommand{\arraystretch}{1.2}
\begin{table*}[h]
\centering
\resizebox{0.6\linewidth}{!}{
\begin{tabular}{l*{4}{c}}
\toprule
Methods & Content-VQA & Spatial-VQA & Stylistic-VQA & \textbf{Average} \\ 
\midrule

\noalign{\smallskip}
\hdashline
\noalign{\smallskip}

SD3.5-Large~\cite{esser2024scaling} & 86.26 {\scriptsize $\pm$ 0.21} & 77.70 {\scriptsize $\pm$ 0.73} & 87.53 {\scriptsize $\pm$ 0.46} & 83.83 {\scriptsize $\pm$ 0.29}\\
+ CADS~\cite{sadat2024cads} & 75.15 {\scriptsize $\pm$ 0.86} & 66.09 {\scriptsize $\pm$ 0.55} & 76.68 {\scriptsize $\pm$ 0.78} & 72.64 {\scriptsize $\pm$ 0.46}\\
\rowcolor{NavyBlue!5}
+ \abbrev & 81.16 {\scriptsize $\pm$ 0.35} & 74.51 {\scriptsize $\pm$ 0.63} & 85.15 {\scriptsize $\pm$ 0.31} & \textbf{80.27} {\scriptsize $\pm$ 0.20}\\

\noalign{\smallskip}
\hdashline
\noalign{\smallskip}

Flux.1-Krea-Dev~\cite{flux1kreadev2025} & 93.28 {\scriptsize $\pm$ 0.23} & 83.28 {\scriptsize $\pm$ 0.39} & 92.93 {\scriptsize $\pm$ 0.10} & 89.83 {\scriptsize $\pm$ 0.15}\\
+ CADS & 89.96 {\scriptsize $\pm$ 0.15} & 78.30 {\scriptsize $\pm$ 0.16} & 87.35 {\scriptsize $\pm$ 0.44} & 85.20 {\scriptsize $\pm$ 0.15}\\
\rowcolor{NavyBlue!5}
+ \abbrev & 89.80 {\scriptsize $\pm$ 0.18} & 80.30 {\scriptsize $\pm$ 0.64} & 91.68 {\scriptsize $\pm$ 0.32} & \textbf{87.26} {\scriptsize $\pm$ 0.23}\\

\noalign{\smallskip}
\hdashline
\noalign{\smallskip}

CogView4~\cite{zheng2024cogview} & 89.33 {\scriptsize $\pm$ 0.41} & 83.16 {\scriptsize $\pm$ 0.43} & 89.30 {\scriptsize $\pm$ 0.44} & 87.27 {\scriptsize $\pm$ 0.17} \\
+ CADS & 52.26 {\scriptsize $\pm$ 1.52} & 57.00 {\scriptsize $\pm$ 0.98} & 58.30 {\scriptsize $\pm$ 1.10} & 55.85 {\scriptsize $\pm$ 1.10} \\
\rowcolor{NavyBlue!5}
+ \abbrev & 85.34 {\scriptsize $\pm$ 0.46} & 79.88 {\scriptsize $\pm$ 0.18} & 88.00 {\scriptsize $\pm$ 0.35} & \textbf{84.41} {\scriptsize $\pm$ 0.20}\\

\noalign{\smallskip}
\hdashline
\noalign{\smallskip}

Qwen-Image~\cite{wu2025qwen} & 95.64 {\scriptsize $\pm$ 0.14} & 85.07 {\scriptsize $\pm$ 0.13} & 93.04 {\scriptsize $\pm$ 0.19} & 91.25 {\scriptsize $\pm$ 0.07} \\
+ CADS & 86.31 {\scriptsize $\pm$ 0.75} & 77.51 {\scriptsize $\pm$ 0.64} & 82.08 {\scriptsize $\pm$ 0.72} & 81.97 {\scriptsize $\pm$ 0.58}\\
\rowcolor{NavyBlue!5}
+ \abbrev & 93.88 {\scriptsize $\pm$ 0.19} & 83.45 {\scriptsize $\pm$ 0.36} & 92.61 {\scriptsize $\pm$ 0.09} & \textbf{89.98} {\scriptsize $\pm$ 0.18}\\
\bottomrule
\end{tabular}
}
\caption{
    \textbf{Additional VQA score with \texttt{Gemma-4-31B-it} on LPD-Bench.}
}
\label{tab:lbp_gemma}
\end{table*}
}

To assess potential judge bias, we additionally evaluate semantic fidelity with another VLM, \texttt{Gemma-4-31B-it}~\cite{team2026gemma}, as an independent VQA judge. The quantitative results are reported in~\cref{tab:lbp_gemma}. We observe the same overall trend as in~\cref{tab:lbp_results}: our method consistently preserve semantic fidelity across all evaluated models. Differences are also consistent across the two judges. For example, CADS exhibits a noticeable degradation on CogView4 under both evaluators, resulting in a similar reduction in the overall average score. These consistent findings indicate that our conclusions are robust to the choice of VQA evaluator, validating the reliability of the adopted VQA-based semantic fidelity metric.

\subsection{Applying \abbrev\ to Autoregressive Models}\label{sec:auto_benchmarking}

\paragraph{\textbf{Comparison between AR and RF Models.}}

To complement our benchmark for evaluating long-prompt fidelity and diversity, we additionally assess several state-of-the-art autoregressive (AR) models, as reported in~\cref{tab:lbp_auto}. Compared with RF-based models, AR models consistently exhibit higher diversity. We attribute this to their inference strategy: once the initial noise is sampled, RF-based models follow a fixed denoising trajectory and cannot introduce new stochasticity. In contrast, AR models resample pixels (or tokens) sequentially, giving each position multiple opportunities to adopt different values. This sequential resampling inherently encourages diverse outputs. These observations further underscore the importance of developing plug-and-play methods that enhance the diversity of RF-based models without sacrificing semantic fidelity.

\vspace{-8pt}
\paragraph{\textbf{Extending \abbrev\ to Autoregressive Models.}} Since our framework only requires access to the conditional distribution $p(\mathbf{X}\mid c)$, it naturally extends to autoregressive models. Specifically, for an autoregressive image generator that produces $k$ tokens $\{x_1, x_2, \ldots, x_k\}$, the conditional likelihood is:
\vspace{-4pt}
\begin{equation*}
    p(x_1, \ldots, x_k \mid c)
    = \prod_{i=1}^{k} p\!\left(x_i \mid x_1, \ldots, x_{i-1}, c\right).
    \vspace{-4pt}
\end{equation*}
By treating the subsequent image tokens $\{x_1, x_2, \ldots, x_k\}$ as a random variable $\mathbf{X}$ and the given prompt token embeddings as $c$, we can directly apply \abbrev\ by sampling prompt embeddings and then generating tokens under the resulting conditioning. We report results on Show-o2 and BLIP3o-NEXT in~\cref{tab:lbp_auto}, demonstrating improved diversity while maintaining semantic fidelity.

{
\renewcommand{\arraystretch}{1.2}
\begin{table*}[h]
\centering
\resizebox{\linewidth}{!}{
\begin{tabular}{l*{8}{c}}
\toprule
\multirow{2}{*}[-0.5ex]{Methods} & \multirow{2}{*}[-0.5ex]{Content-VQA} & \multirow{2}{*}[-0.5ex]{Spatial-VQA} & \multirow{2}{*}[-0.5ex]{Stylistic-VQA} & \multicolumn{5}{c}{\textbf{Average}} \\ \cmidrule(lr){5-9}
& & & & VQA & AS & VS-Incep. & VS-DINO & Overall $\downarrow$ \\
\midrule
% \rowcolor{Gray!5}
Janus-Pro-7B{\small $^{384}$}~\cite{chen2025janus} & 79.23 {\scriptsize $\pm$ 0.59} & 67.82 {\scriptsize $\pm$ 0.17} & 79.87 {\scriptsize $\pm$ 0.28} & 75.64 {\scriptsize $\pm$ 0.27} & 5.59 {\scriptsize $\pm$ 0.01} & 2.27 {\scriptsize $\pm$ 0.26} & 2.11 {\scriptsize $\pm$ 0.42} & - \\

\noalign{\smallskip}
\hdashline
\noalign{\smallskip}

% \rowcolor{Gray!5}
NextStep-1{\small $^{512}$}~\cite{team2025nextstep} & 88.35 {\scriptsize $\pm$ 0.26} & 74.81 {\scriptsize $\pm$ 0.22} & 83.34 {\scriptsize $\pm$ 0.27} & 82.17 {\scriptsize $\pm$ 0.15} & 6.01 {\scriptsize $\pm$ 0.05} & 2.05 {\scriptsize $\pm$ 0.28} & 1.88 {\scriptsize $\pm$ 0.40} & - \\

\noalign{\smallskip}
\hdashline
\noalign{\smallskip}

% \rowcolor{Gray!5}
Show-o2~\cite{xie2025showo} & 86.16 {\scriptsize $\pm$ 0.49} & 75.03 {\scriptsize $\pm$ 0.33} & 82.00 {\scriptsize $\pm$ 0.17} & 81.06 {\scriptsize $\pm$ 0.26} & 6.05 {\scriptsize $\pm$ 0.01} & 1.90 {\scriptsize $\pm$ 0.20} & 1.56 {\scriptsize $\pm$ 0.26} & - \\
\rowcolor{NavyBlue!5}
+ \abbrev & 84.20 {\scriptsize $\pm$ 0.21} & 73.74 {\scriptsize $\pm$ 0.22} & 81.31 {\scriptsize $\pm$ 0.20} & 79.75 {\scriptsize $\pm$ 0.14} & 5.99 {\scriptsize $\pm$ 0.01} & \textbf{1.96}* {\scriptsize $\pm$ 0.20} & \textbf{1.66}* {\scriptsize $\pm$ 0.30} & \textbf{0.46} \\

\noalign{\smallskip}
\hdashline
\noalign{\smallskip}

% \rowcolor{Gray!5}
BLIP3o-NEXT~\cite{chen2025blip3o} & 85.39 {\scriptsize $\pm$ 0.20} & 73.74 {\scriptsize $\pm$ 0.23} & 81.99 {\scriptsize $\pm$ 0.21} & 80.37 {\scriptsize $\pm$ 0.11} & 6.41 {\scriptsize $\pm$ 0.01} & 1.92 {\scriptsize $\pm$ 0.23} & 1.59 {\scriptsize $\pm$ 0.28} & - \\
\rowcolor{NavyBlue!5}
+ \abbrev & 83.38 {\scriptsize $\pm$ 0.37} & 72.93 {\scriptsize $\pm$ 0.50} & 81.51 {\scriptsize $\pm$ 0.09} & 79.27 {\scriptsize $\pm$ 0.24} & 6.36 {\scriptsize $\pm$ 0.01} & \textbf{2.00}* {\scriptsize $\pm$ 0.24} & \textbf{1.69}* {\scriptsize $\pm$ 0.32} & \textbf{0.34} \\
\bottomrule
\end{tabular}
}
\caption{
    \textbf{Additional results of autoregressive-based models on LPD-Bench.} The default image resolution is $1024{\times}1024$, while methods marked with a superscript indicate their supported resolution. (*: $p\text{-value} \approx 10^{-5} \le 10^{-3}$, with Welch's t-test across prompts.)
}
\label{tab:lbp_auto}
\end{table*}
}

\section{Illustration and Algorithm of \abbrev}\label{sec:ill_algo}

\begin{figure}[h]
    \centering
    \includegraphics[width=0.8\linewidth]{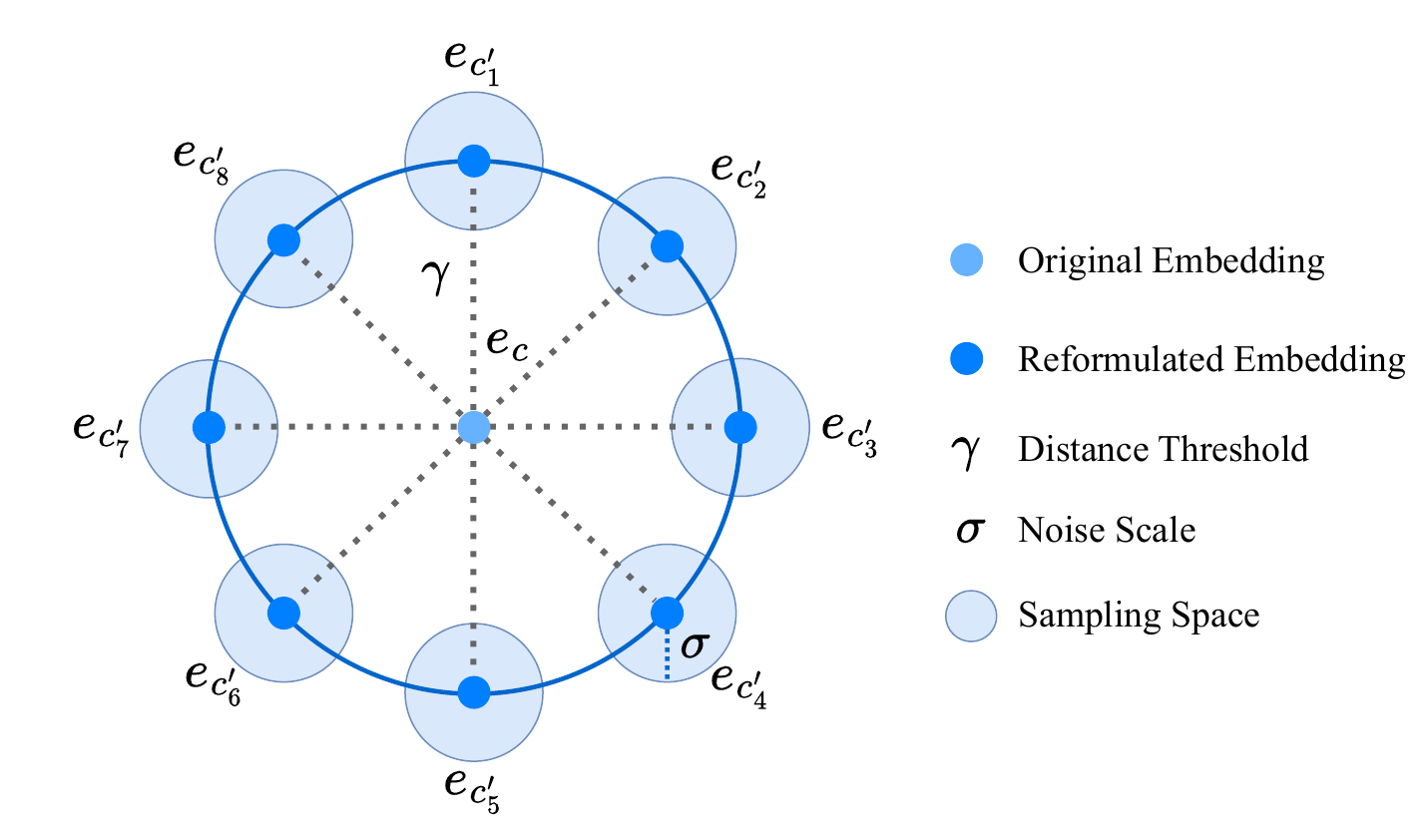}
    \caption{\textbf{Illustration of \abbrev\ for $n=8$ in a 2D embedding space.}}
    \label{fig:prompt-mog-example}
\end{figure}

We provide a visualization of the feasible region in~\cref{fig:prompt-mog-example}. Given a distance threshold $\gamma$, the set of embeddings forms a spherical surface. We then apply a regular simplex construction to place $n$ reformulated embeddings uniformly on the spherical surface, maximizing their pairwise separation. These embeddings serve as the centers of the Gaussian components, forming a Mixture-of-Gaussians (MoG) used in the sampling process.

In~\cref{alg:prompt-mog}, we present the complete algorithm of our approach. The only additional step required during inference is sampling a prompt embedding from the constructed MoG. Once sampled, this embedding is reused throughout the generation process, which follows the original rectified flow formulation without modification.

\begin{algorithm}[h]
    \DontPrintSemicolon
    \SetAlgoLined
    \caption{Generation with \abbrev}
    \label{alg:prompt-mog}
    \KwIn{prompt $c$; text encoder $E(\cdot)$; distance threshold $\gamma_{\text{euc}}$; noise scale $\sigma$; number of modes $n$; model $\mathbf{v}_\theta(\cdot)$; total timesteps $t_{\tmax}$}
    \KwOut{Generated image $\mathbf{x}_0$}
    \BlankLine
    $\boldsymbol{e}_c \gets E(c)$ \tcp*{Embedding from original prompt}
    $\{\boldsymbol{u}_i\}_{i=1}^n \gets$ vertices of a simplex on the $d$-sphere\;
    $\boldsymbol{e}_{c'_i} \gets \boldsymbol{e}_c + \gamma_{\text{euc}}\, \boldsymbol{u}_i, \quad i = 1, \dots, n$ \;
    $\boldsymbol{e}_{c'} \sim \tfrac{1}{n}\sum_i \mathcal{N}(\boldsymbol{e}_{c'_i}, \sigma^2 \mathbf{I})$ \tcp*{Sampled from MoG}
    $\mathbf{x}_{t_{\tmax}} \sim \mathcal{N}(\mathbf{0}, \mathbf{I})$\;
    \For(\tcp*[f]{Denoising $\mathbf{x}_{t_k}$}){$k = \tmax, \dots, 1$}{
        $\mathbf{x}_{t_{k-1}} \gets \mathbf{x}_{t_k} + (t_{k-1}-t_k)\,\mathbf{v}_\theta(\mathbf{x}_{t_k}, t_k, \boldsymbol{e}_{c'})$\;
    }
    \Return $\mathbf{x}_0$\;
\end{algorithm}

\section{Limitation}\label{sec:limitation}

\begin{figure}[h]
    \centering
    \includegraphics[width=\linewidth]{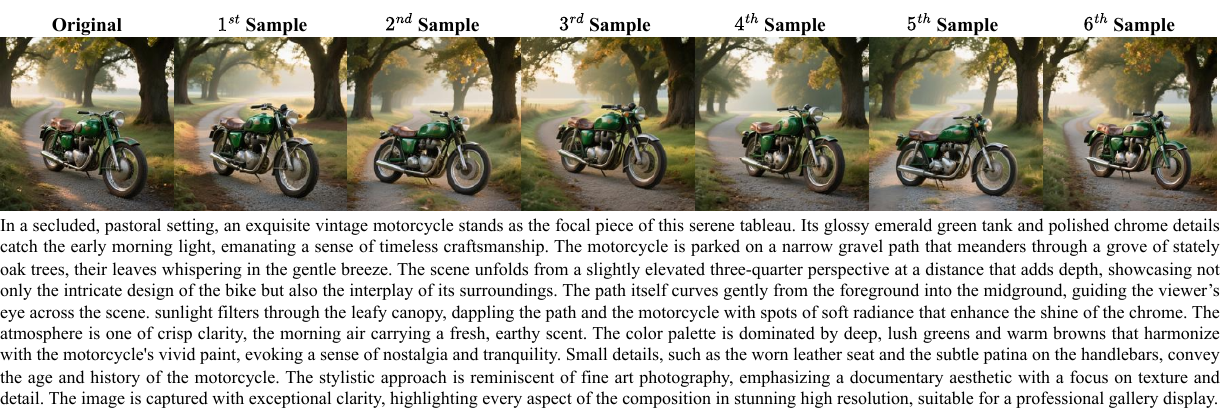}
    \caption{\textbf{Illustration of similar results across sampling trials.} Images are generated by Qwen-Image~\cite{wu2025qwen}. The leftmost column shows the original output, while the remaining columns show \abbrev\ under different sampling trials.}
    \label{fig:limitation}
\end{figure}

Since our approach constrains sampling points to lie near a hypersphere, it generally preserves semantic fidelity. However, our analysis assumes that any point deviating from the original prompt embedding manifold yields an entropy gain relative to the original prompt embedding. As shown in~\cref{fig:limitation}, across six sampling attempts, the generated outputs remain similar because the sampled points introduce little to no entropy gain. 

To address this issue, users can either \textbf{(1)} increase the number of sampling trials to better locate points that provide entropy gain, or \textbf{(2)} incorporate an explicit truncation mechanism in the embedding space that filters out redundant points (\eg, those with negligible entropy gain). Compared with \textbf{(1)}, \textbf{(2)} improves sampling efficiency but requires additional design, which we do not study in this work. We leave the study of efficient truncation strategies and the broader question of maximizing diversity under a fixed sampling budget as future work.

\section{Additional Visualization Comparison}\label{sec:additional_comp}

\begin{figure*}[h]
    \centering
    \includegraphics[width=\linewidth]{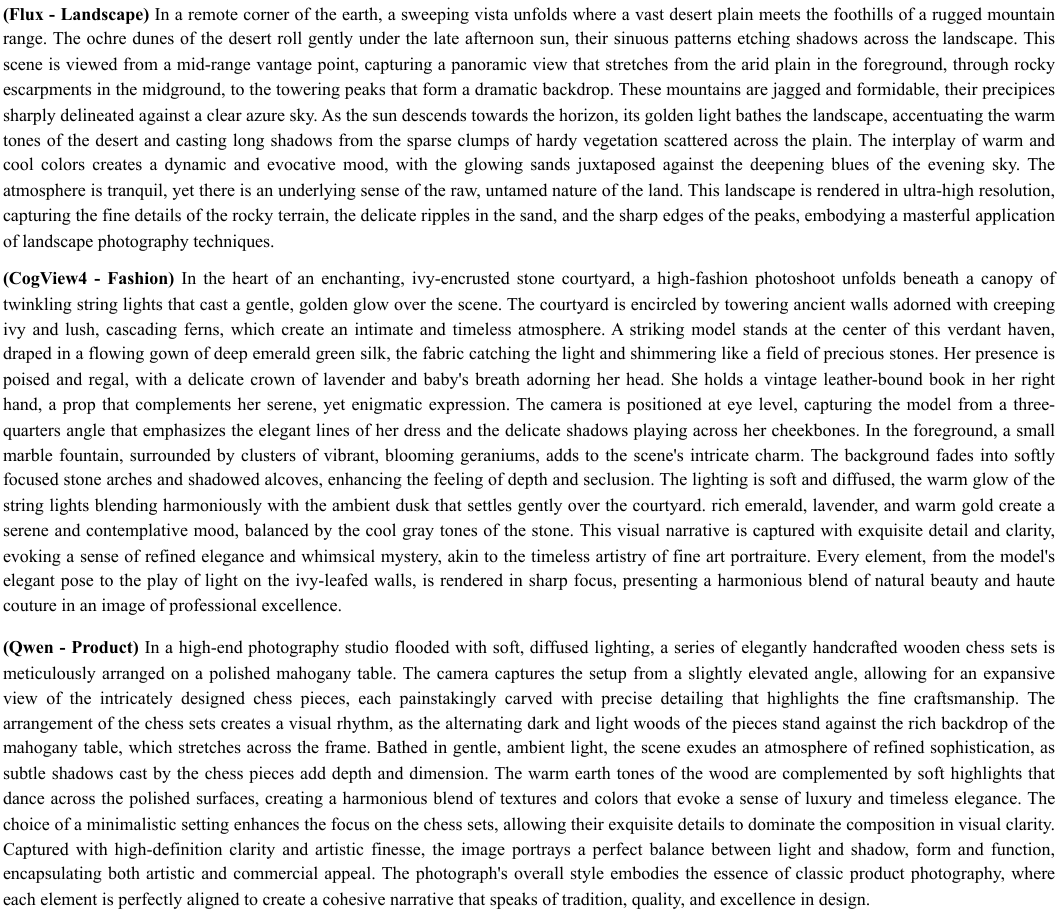}
    \caption{\textbf{Illustration of the full prompts used in \cref{fig:qual}.} Themes are indicated at the beginning of each prompt in \textbf{bold}.}
    \label{fig:full_prompt}
\end{figure*}

The full prompts used in \cref{fig:qual} are shown in \cref{fig:full_prompt}. We provide additional qualitative comparisons between \abbrev\ \textbf{(ours)}, CADS~\cite{sadat2024cads}, and DiverseFlow~\cite{morshed2025diverseflow} on SD3.5-Large~\cite{esser2024scaling} (\cref{fig:sd3_extra}), Flux.1-Krea-Dev~\cite{flux1kreadev2025} (\cref{fig:flux_extra}), CogView4~\cite{zheng2024cogview} (\cref{fig:cogview_extra}), and Qwen-Image~\cite{wu2025qwen} (\cref{fig:qwen_extra}). For each model, we include 3 comparisons across different themes. Consistent with~\cref{fig:qual}, CADS frequently produces checkerboard artifacts or deviates significantly from the prompt. DiverseFlow often yields over-optimized outputs with distorted or unappealing styles; although it performs slightly better on CogView4, oversaturated colors remain evident. Overall, our method consistently generates diverse outputs while preserving alignment with the original prompt.

\begin{figure*}[h]
    \centering
    \begin{subfigure}{\linewidth}
        \centering
        \includegraphics[width=\linewidth]{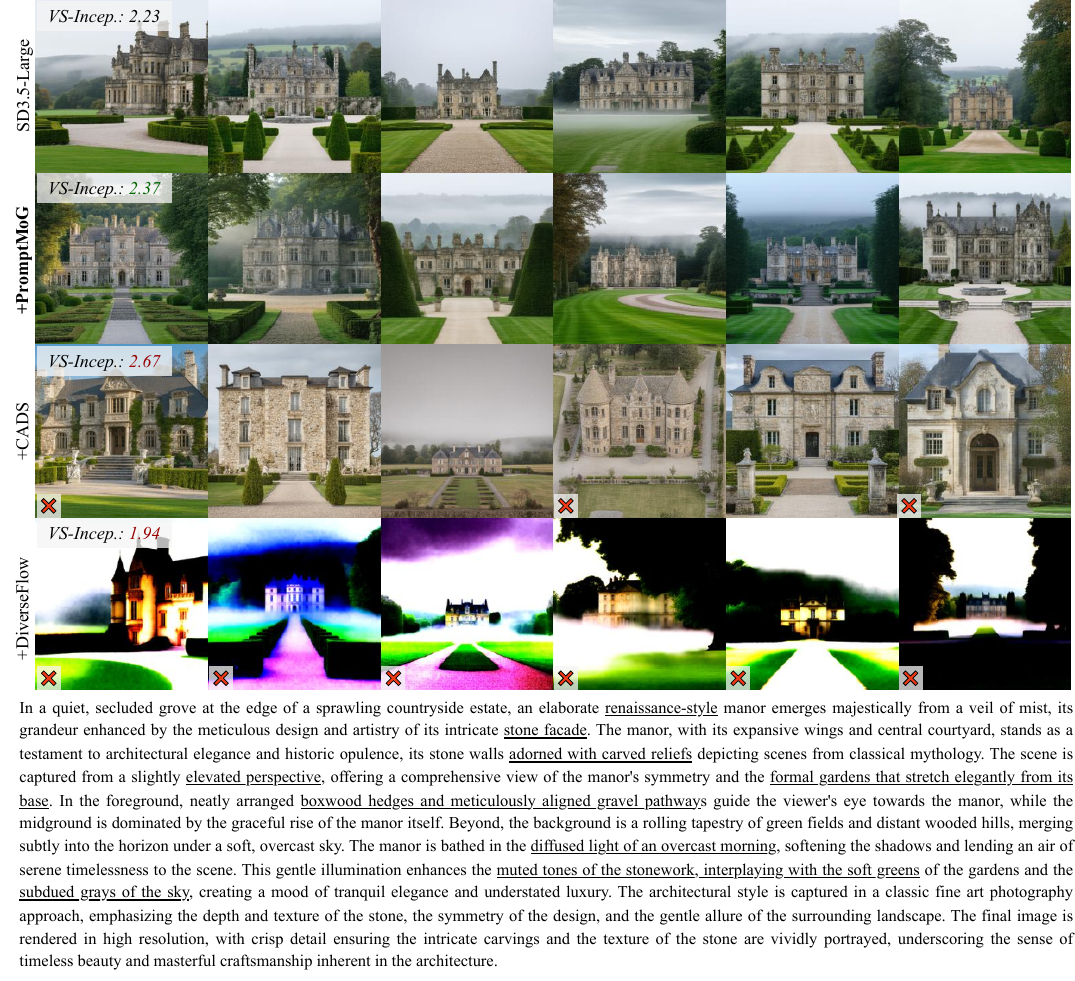}
        \vspace{1pt}
        \caption{\textbf{Theme: \textit{Architecture}}}
    \end{subfigure}
\end{figure*}
\begin{figure*}[h]
    \ContinuedFloat
    \centering
    \begin{subfigure}{\linewidth}
        \centering
        \includegraphics[width=\linewidth]{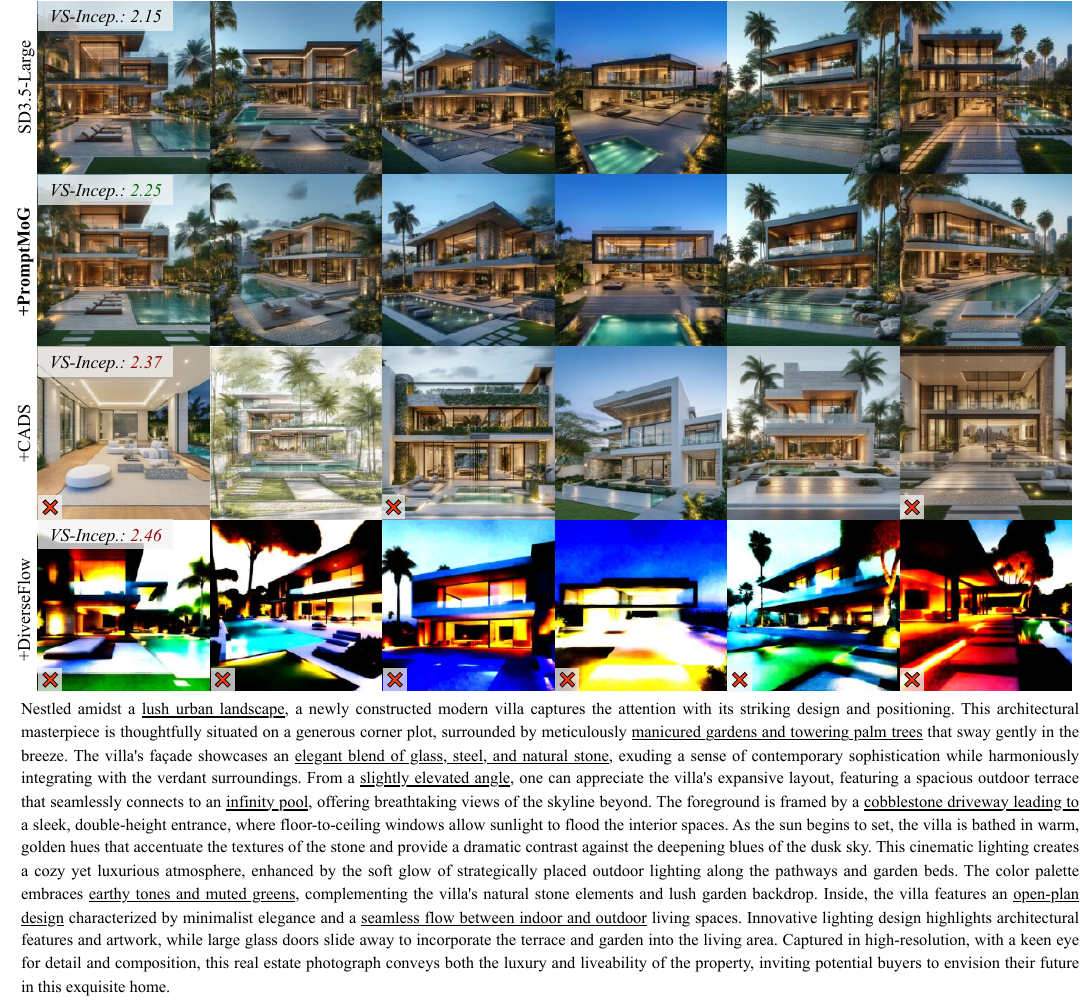}
        \vspace{1pt}
        \caption{\textbf{Theme: \textit{Real Estate}}}
    \end{subfigure}
\end{figure*}
\begin{figure*}[h]
    \ContinuedFloat
    \centering
    \begin{subfigure}{\linewidth}
        \centering
        \includegraphics[width=\linewidth]{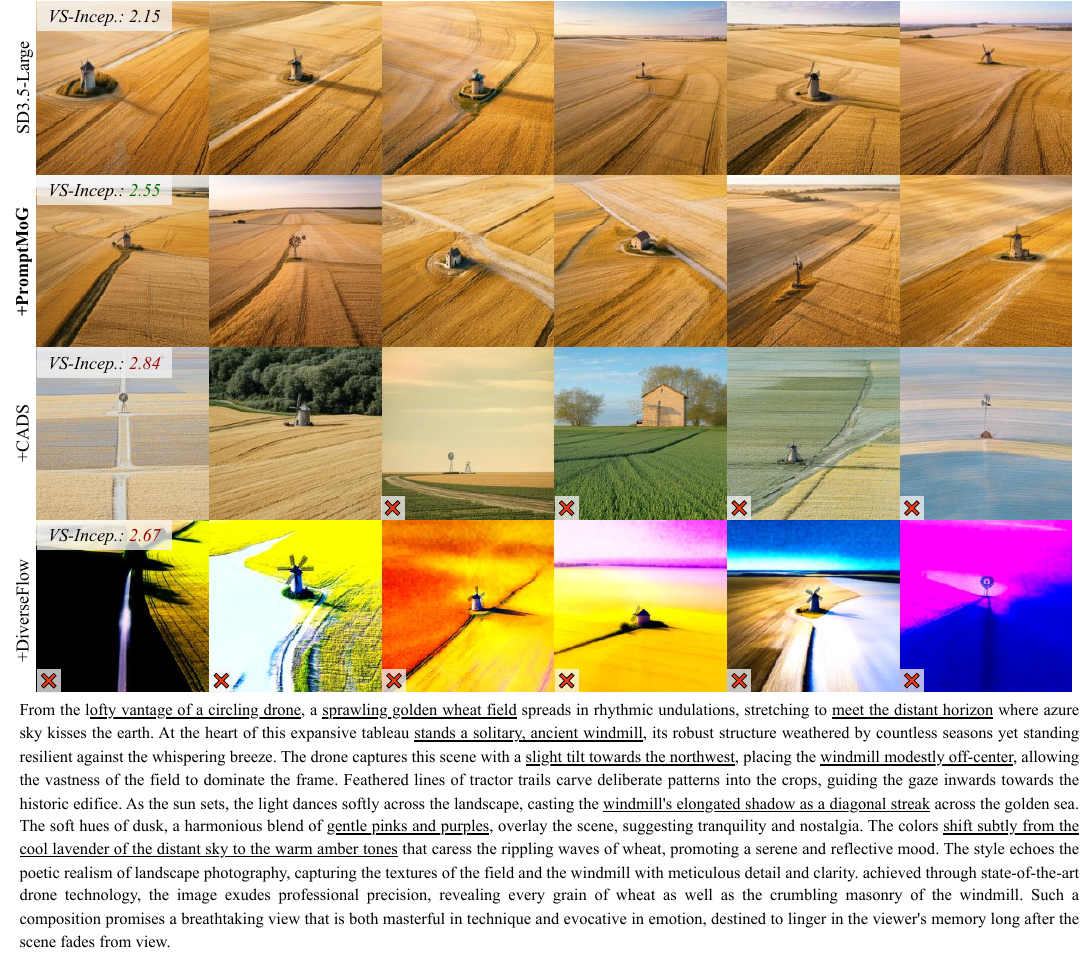}
        \vspace{1pt}
        \caption{\textbf{Theme: \textit{Aerial}}}
    \end{subfigure}
    \vspace{1pt}
    \caption{\textbf{Qualitative comparison with SD3.5-Large~\cite{esser2024scaling}.} Vendi Scores are shown at the top left, and failed outputs are marked at the bottom left. Themes are indicated in each sub-caption.}
    \label{fig:sd3_extra}
\end{figure*}

\begin{figure*}[h]
    \centering
    \begin{subfigure}{\linewidth}
        \centering
        \includegraphics[width=\linewidth]{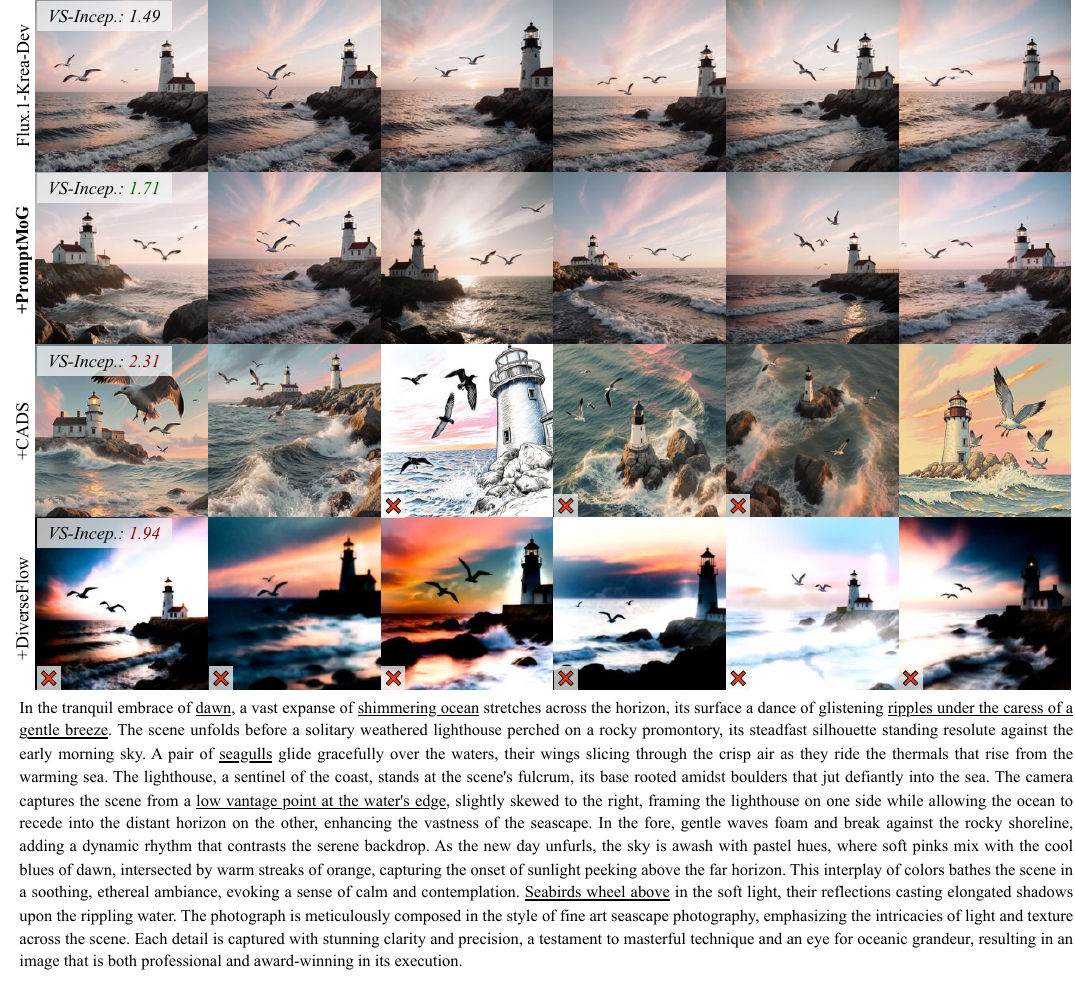}
        \vspace{1pt}
        \caption{\textbf{Theme: \textit{Seascape}}}
    \end{subfigure}
\end{figure*}
\begin{figure*}[h]
    \ContinuedFloat
    \centering
    \begin{subfigure}{\linewidth}
        \centering
        \includegraphics[width=\linewidth]{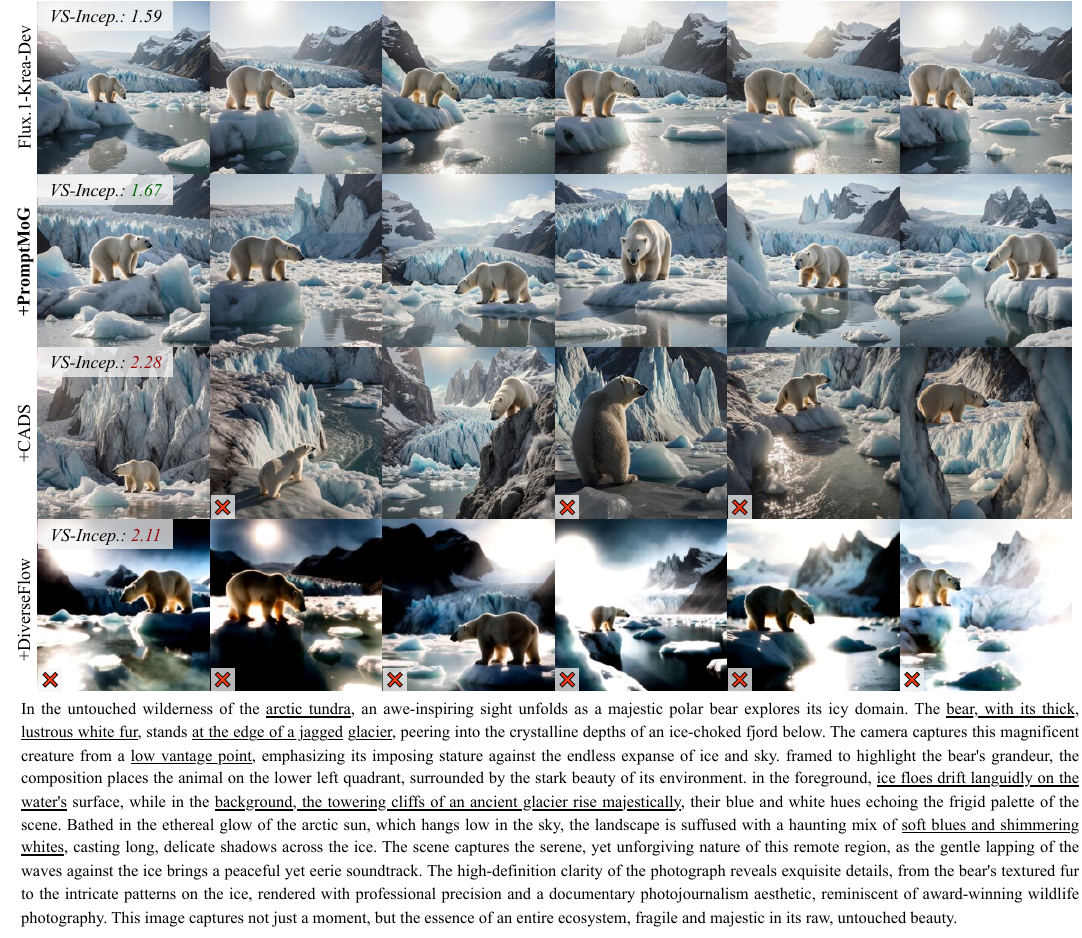}
        \vspace{1pt}
        \caption{\textbf{Theme: \textit{Wildlife}}}
    \end{subfigure}
\end{figure*}
\begin{figure*}[h]
    \ContinuedFloat
    \centering
    \begin{subfigure}{\linewidth}
        \centering
        \includegraphics[width=\linewidth]{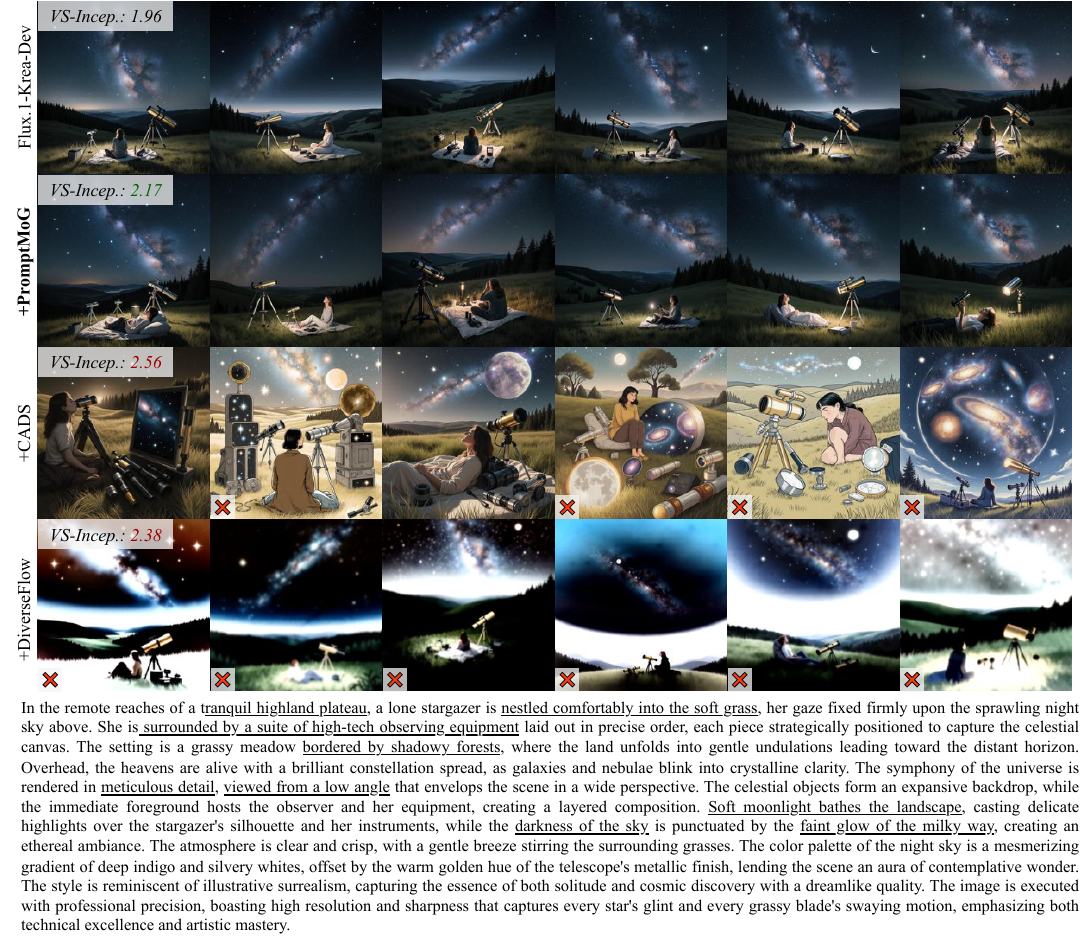}
        \vspace{1pt}
        \caption{\textbf{Theme: \textit{Astrophotography}}}
    \end{subfigure}
    \vspace{1pt}
    \caption{\textbf{Qualitative comparison with Flux.1-Krea-Dev~\cite{flux1kreadev2025}.} Vendi Scores are shown at the top left, and failed outputs are marked at the bottom left. Themes are indicated in each sub-caption.}
    \label{fig:flux_extra}
\end{figure*}

\begin{figure*}[h]
    \centering
    \begin{subfigure}{\linewidth}
        \centering
        \includegraphics[width=\linewidth]{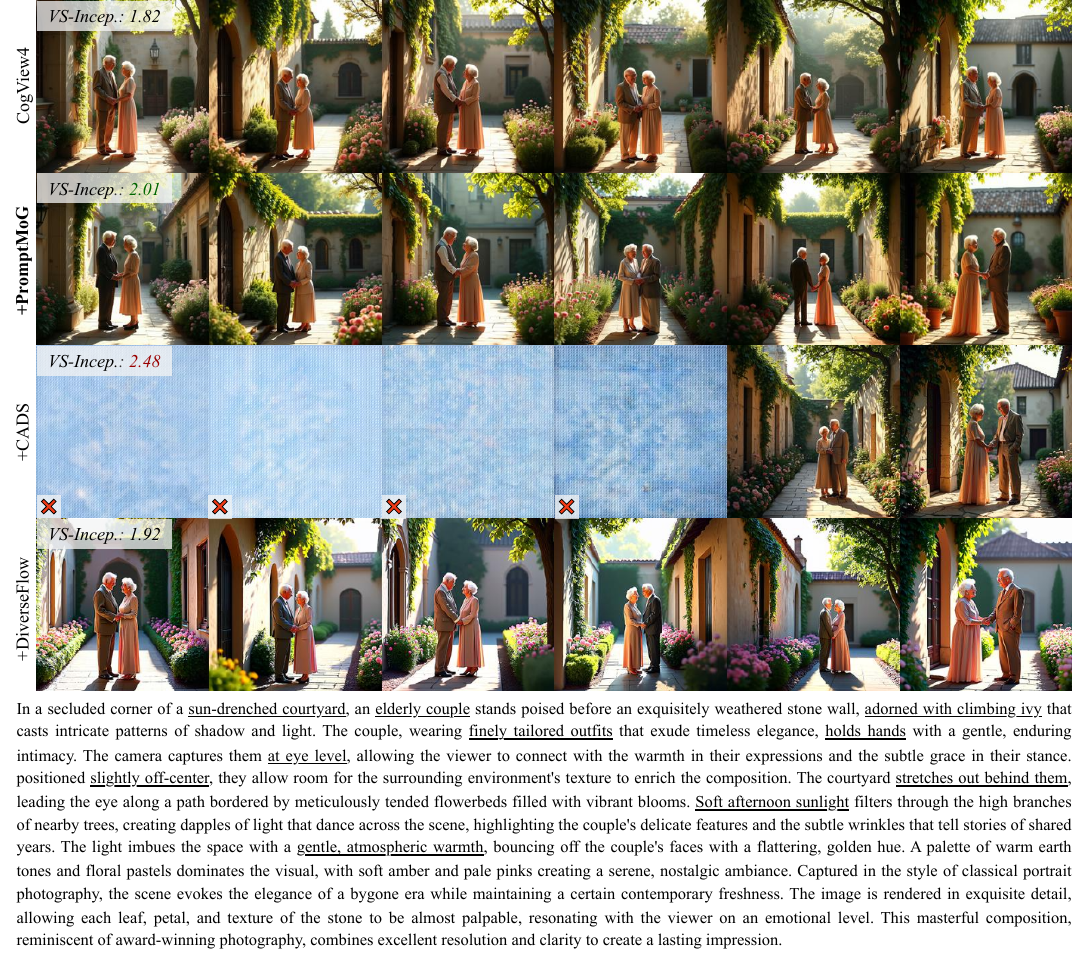}
        \vspace{1pt}
        \caption{\textbf{Theme: \textit{Portrait}}}
    \end{subfigure}
\end{figure*}
\begin{figure*}[h]
    \ContinuedFloat
    \centering
    \begin{subfigure}{\linewidth}
        \centering
        \includegraphics[width=\linewidth]{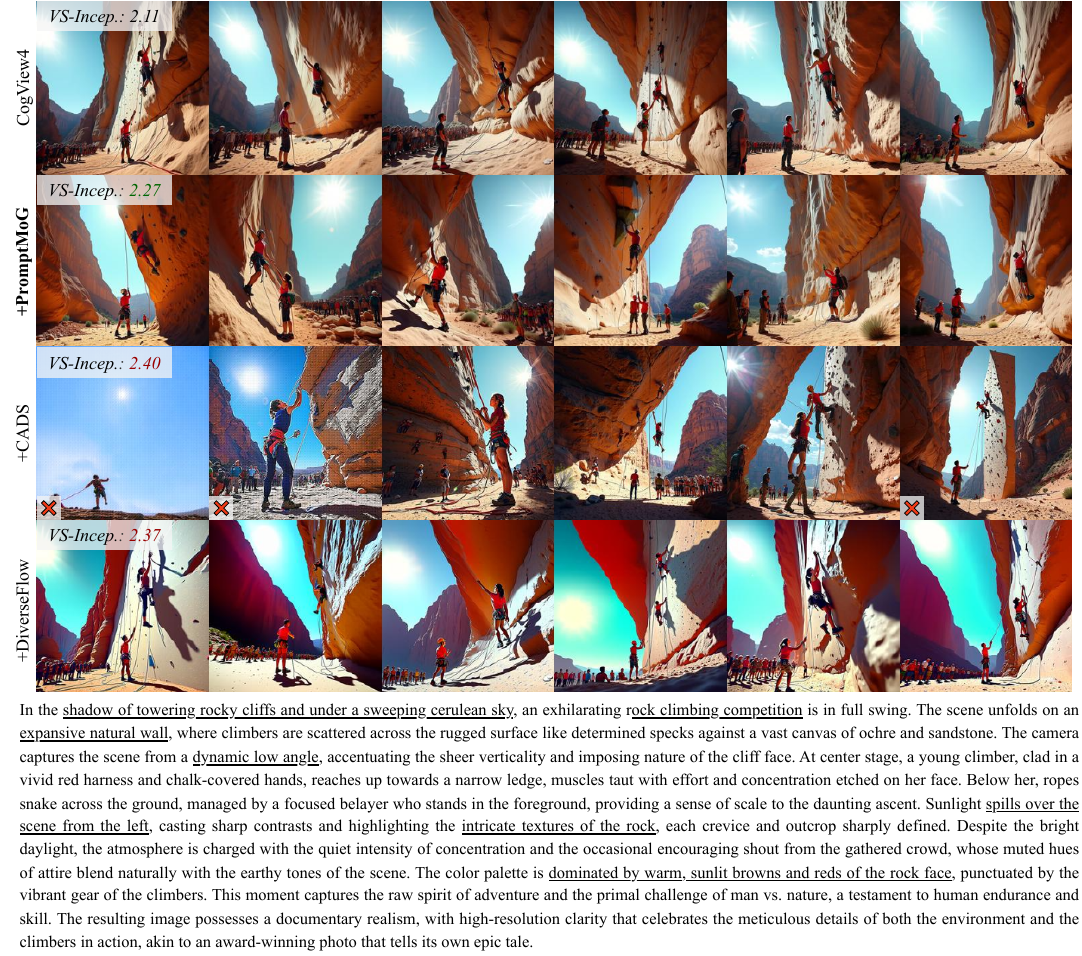}
        \vspace{1pt}
        \caption{\textbf{Theme: \textit{Sport}}}
    \end{subfigure}
\end{figure*}
\begin{figure*}[h]
    \ContinuedFloat
    \centering
    \begin{subfigure}{\linewidth}
        \centering
        \includegraphics[width=\linewidth]{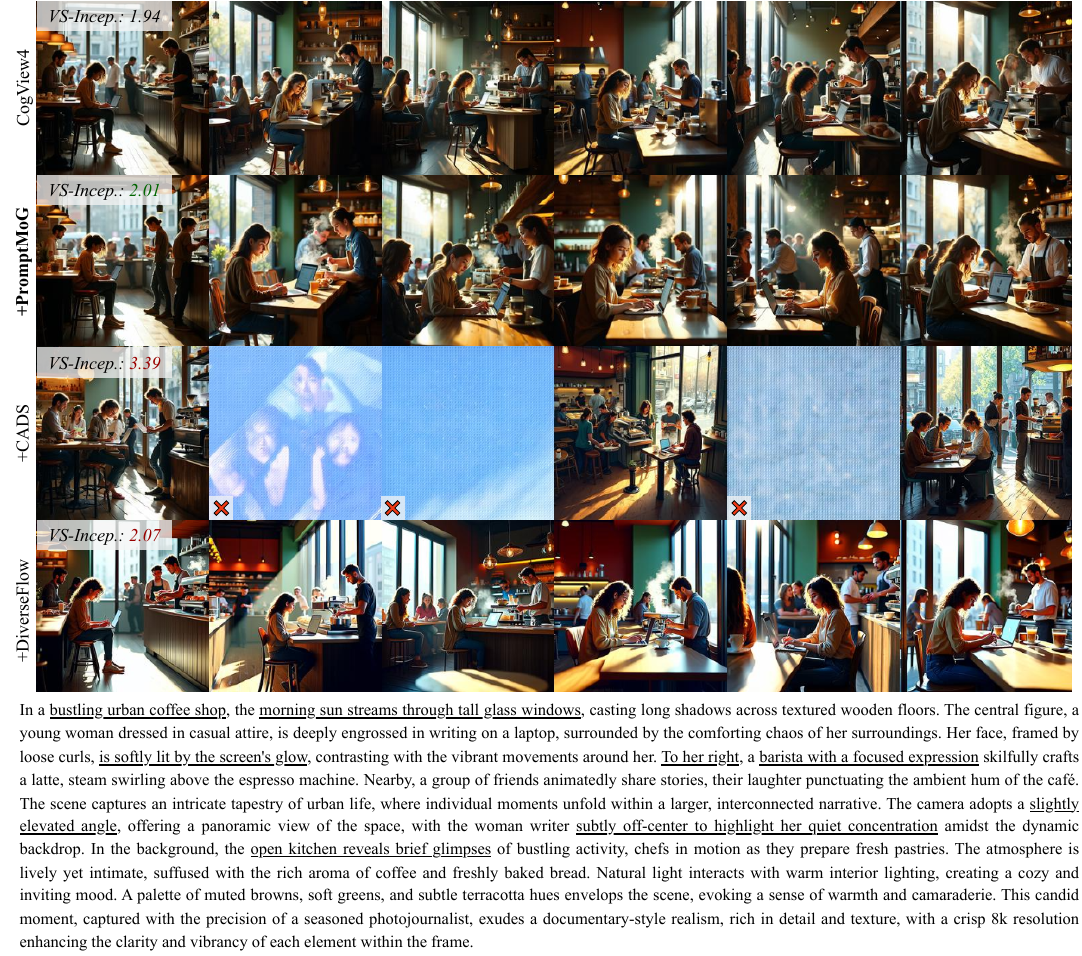}
        \vspace{1pt}
        \caption{\textbf{Theme: \textit{Candid}}}
    \end{subfigure}
    \vspace{1pt}
    \caption{\textbf{Qualitative comparison with CogView4~\cite{zheng2024cogview}.} Vendi Scores are shown at the top left, and failed outputs are marked at the bottom left. Themes are indicated in each sub-caption.}
    \label{fig:cogview_extra}
\end{figure*}

\begin{figure*}[h]
    \centering
    \begin{subfigure}{\linewidth}
        \centering
        \includegraphics[width=\linewidth]{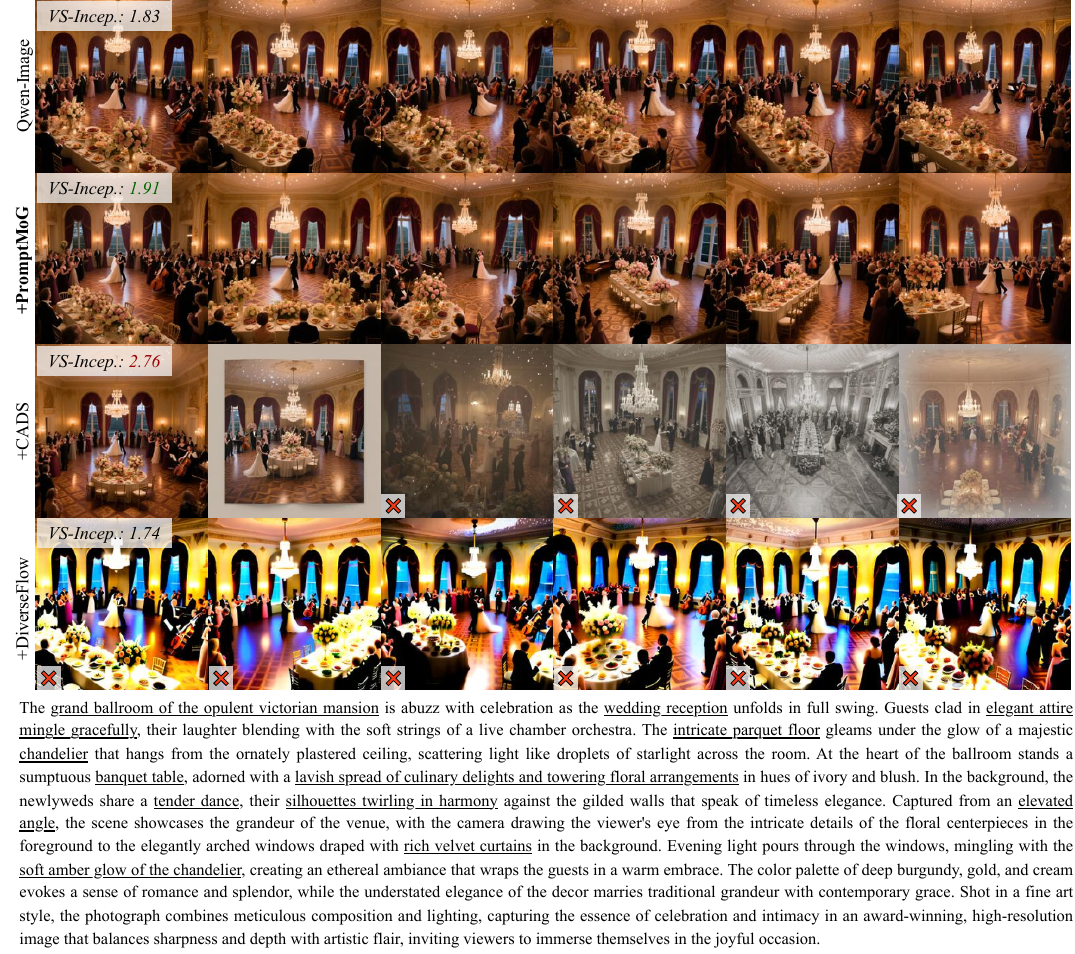}
        \vspace{1pt}
        \caption{\textbf{Theme: \textit{Event}}}
    \end{subfigure}
\end{figure*}
\begin{figure*}[h]
    \ContinuedFloat
    \centering
    \begin{subfigure}{\linewidth}
        \centering
        \includegraphics[width=\linewidth]{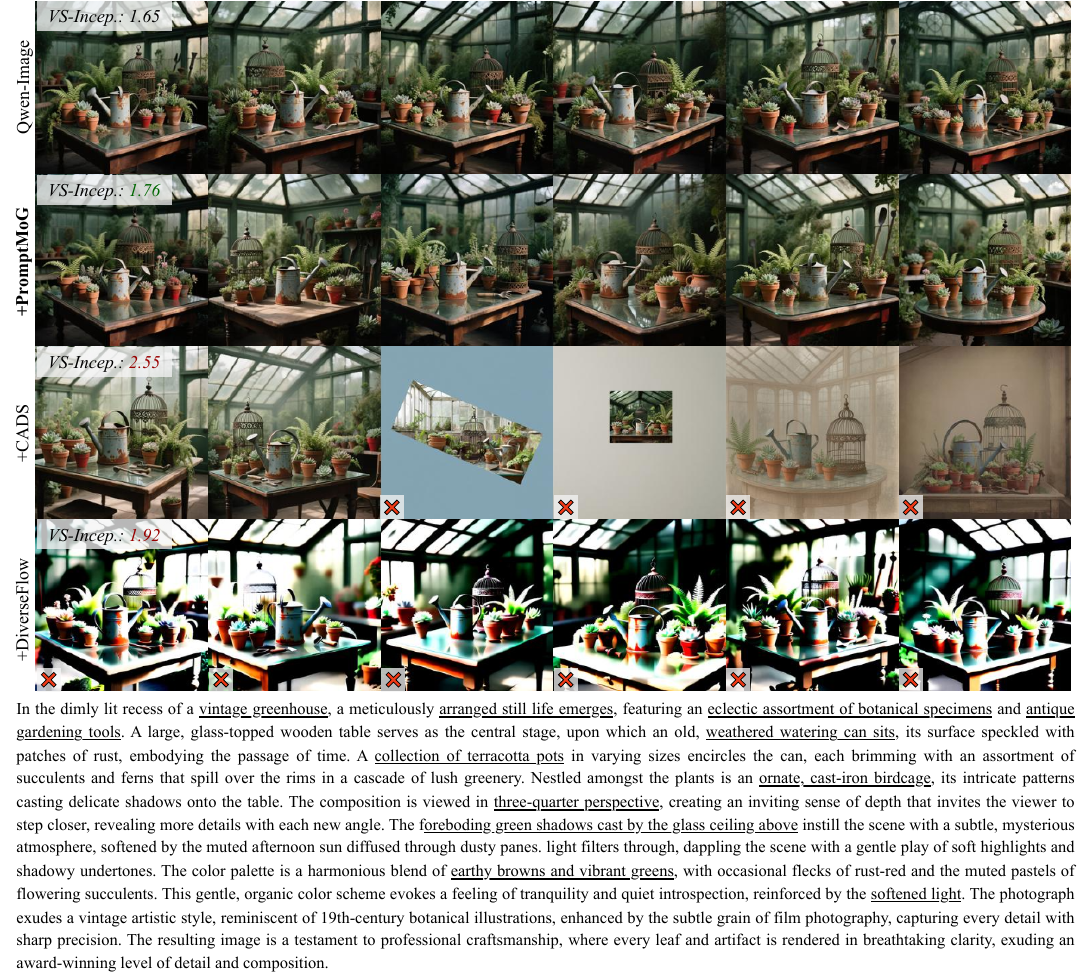}
        \vspace{1pt}
        \caption{\textbf{Theme: \textit{Still Life}}}
    \end{subfigure}
\end{figure*}
\begin{figure*}[h]
    \ContinuedFloat
    \centering
    \begin{subfigure}{\linewidth}
        \centering
        \includegraphics[width=\linewidth]{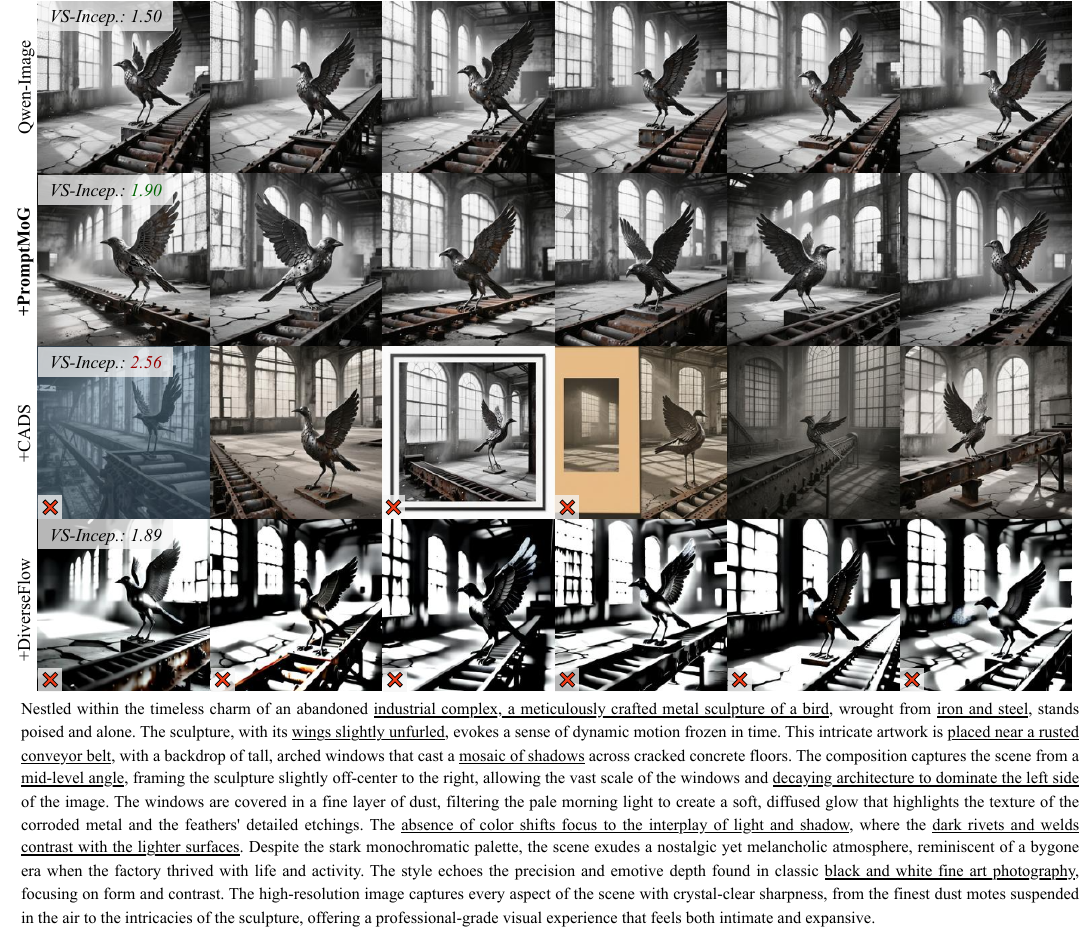}
        \vspace{1pt}
        \caption{\textbf{Theme: \textit{Black and White}}}
    \end{subfigure}
    \vspace{1pt}
    \caption{\textbf{Qualitative comparison with Qwen-Image~\cite{wu2025qwen}.} Vendi Scores are shown at the top left, and failed outputs are marked at the bottom left. Themes are indicated in each sub-caption.}
    \label{fig:qwen_extra}
\end{figure*}

\end{appendices}

\end{document}